\documentclass[10pt,twocolumn,letterpaper]{article}

\usepackage{3dv}
\usepackage{times}
\usepackage{epsfig}
\usepackage{graphicx}
\usepackage{amsmath}
\usepackage{amssymb}
\usepackage{wasysym}
\usepackage[percent]{overpic}
\usepackage{tablefootnote}
\usepackage{wrapfig}
\usepackage[table]{xcolor}
\definecolor{LightCyan}{rgb}{0.88,1,1}
\definecolor{LightYellow}{rgb}{1,1,0.7}
\usepackage{verbatim}
\usepackage{url}
\usepackage[pagebackref=true,breaklinks=true,letterpaper=true,colorlinks,bookmarks=false]{hyperref}

\threedvfinalcopy 

\def\threedvPaperID{****} 

\begin{document}

\title{Learning monocular depth estimation with unsupervised trinocular assumptions}

\author{Matteo Poggi, Fabio Tosi, Stefano Mattoccia\\
University of Bologna, Department of Computer Science and Engineering\\
Viale del Risorgimento 2, Bologna, Italy\\
{\tt\small \{m.poggi, fabio.tosi5, stefano.mattoccia\}@unibo.it}
}

\maketitle
\thispagestyle{empty}

\begin{abstract}
Obtaining accurate depth measurements out of a single image represents a fascinating solution to 3D sensing. CNNs led to considerable improvements in this field, and recent trends replaced the need for ground-truth labels with geometry-guided image reconstruction signals enabling unsupervised training. Currently, for this purpose, state-of-the-art techniques rely on images acquired with a binocular stereo rig to predict inverse depth (i.e., disparity) according to the aforementioned supervision principle. However, these methods suffer from well-known problems near occlusions, left image border, etc inherited from the stereo setup.
Therefore, in this paper, we tackle these issues by moving to a trinocular domain for training. Assuming the central image as the reference, we train a CNN to infer disparity representations pairing such image with frames on its left and right side. This strategy allows obtaining depth maps not affected by typical stereo artifacts. Moreover, being trinocular datasets seldom available, we introduce a novel interleaved training procedure enabling to enforce the trinocular assumption outlined from current binocular datasets.
Exhaustive experimental results on the KITTI dataset confirm that our proposal outperforms state-of-the-art methods for unsupervised monocular depth estimation trained on binocular stereo pairs as well as any known methods relying on other cues.
\end{abstract}

\begin{figure}
    \centering
    \begin{overpic}[width=0.42\textwidth]{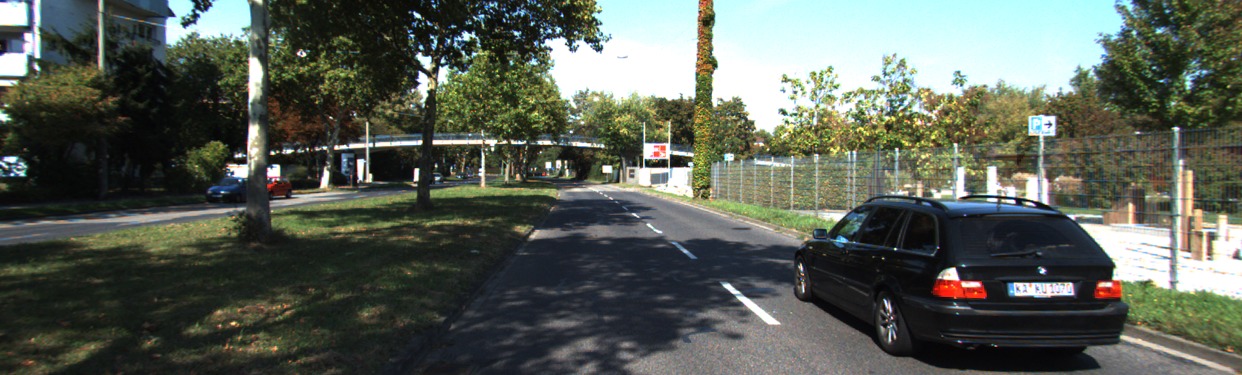}
    \put (2,4) {$\displaystyle\textcolor{white}{\textbf{(a)}}$}
    \end{overpic} \\
    \begin{overpic}[width=0.42\textwidth]{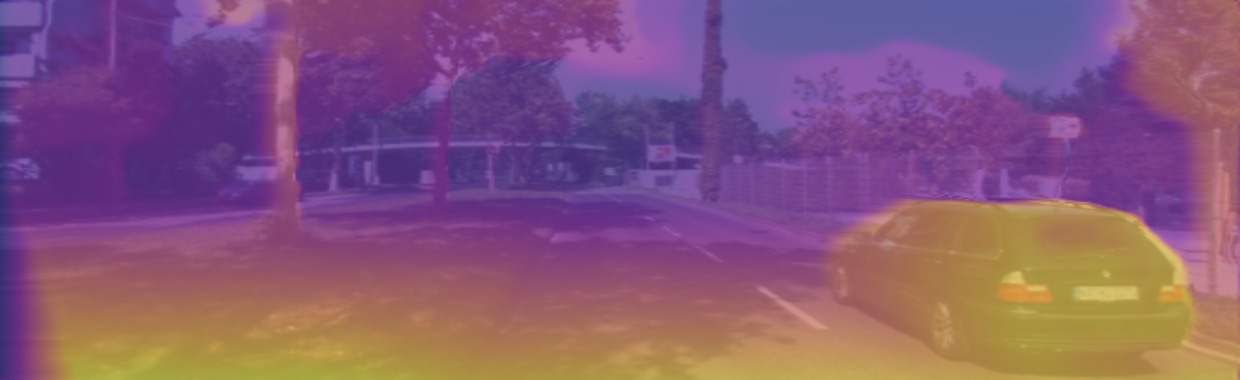}
    \put (2,4) {$\displaystyle\textcolor{white}{\textbf{(b)}}$}
    \put(96,4){\color{white}\vector(1,0){4}}
    \put(100,4){\color{white}\vector(-1,0){4}}
    \end{overpic} \\
    \begin{overpic}[width=0.42\textwidth]{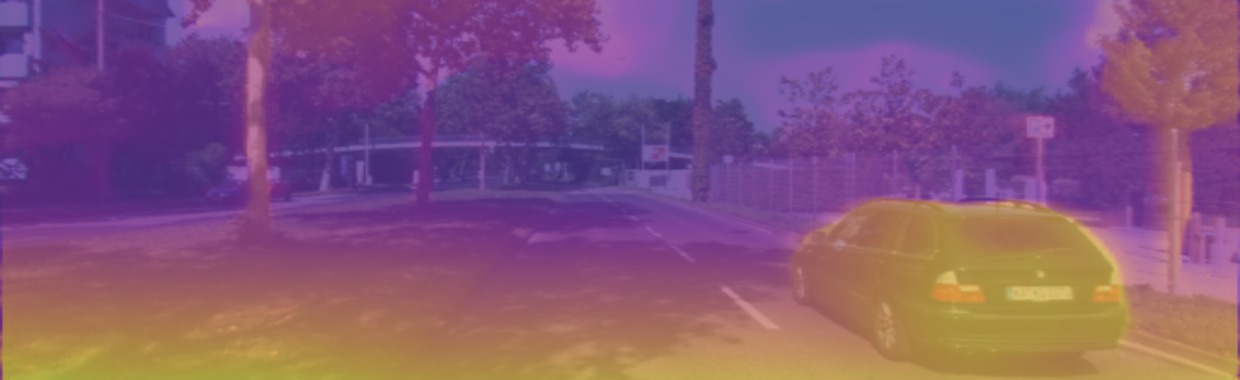}
    \put (2,4) {$\displaystyle\textcolor{white}{\textbf{(c)}}$}
    \put(91,4){\color{white}\vector(1,0){9}}
    \put(100,4){\color{white}\vector(-1,0){9}}    
    \end{overpic} \\
    \begin{overpic}[width=0.42\textwidth]{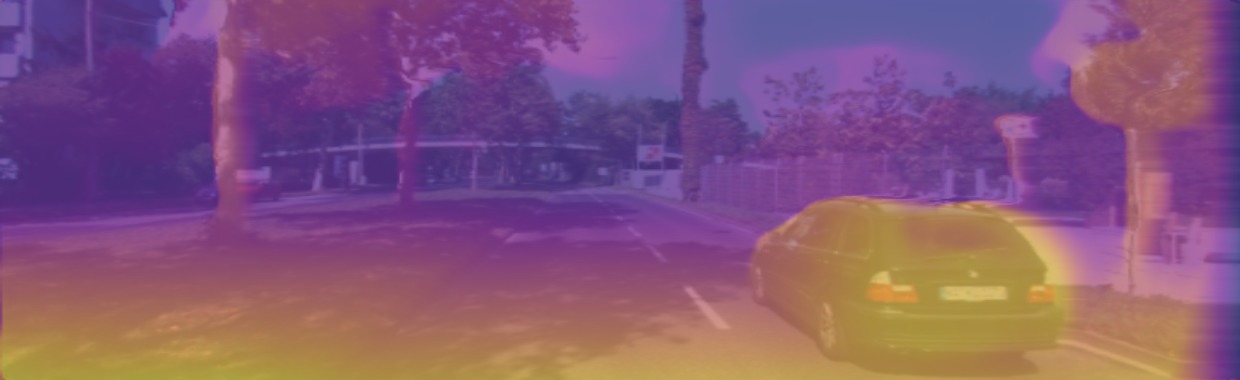}
    \put (2,4) {$\displaystyle\textcolor{white}{\textbf{(d)}}$}
    \put(86,4){\color{white}\vector(1,0){14}}
    \put(100,4){\color{white}\vector(-1,0){14}}
    \end{overpic} \\
    \caption{Overview of 3Net. a) Given a single reference image from KITTI 2015 training set \cite{KITTI_2015}, our network learns depth representations according to two additional points of view on left (b) and right (d) of the input (a), enabling to infer a more accurate depth map (c). White arrows highlight the different points of view.}
    \label{fig:abstract}
\end{figure}
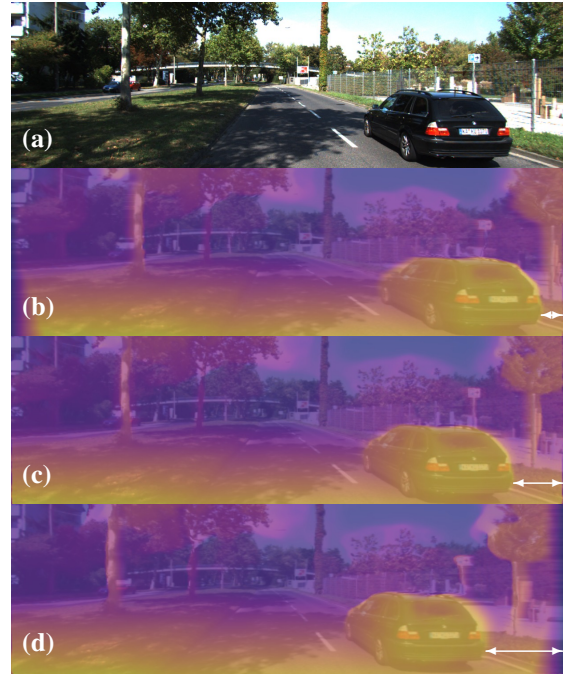

\section{Introduction}

Depth plays a crucial role in many computer vision applications and active 3D sensors are becoming very popular. Nonetheless, such sensors may have severe shortcomings. For instance, the Kinect 1 is not suited at all for outdoor environments flooded by sunlight. Moreover, such sensor allows only for close range depth measurements. On the other hand, a popular active depth sensor perfectly suited for outdoor environments is LIDAR (e.g., Velodyne). However, sensors based on such technology are typically expensive and often cumbersome for some practical applications.
Thus, inferring depth with passive sensors based on standard imaging technology would be highly desirable being cheap, lightweight and suited for indoor and outdoor environments. In this context, acquiring images from different viewpoints allows inferring depth exploiting geometry constraints. On the other hand, estimating depth from a single image is indeed an ill-posed problem. Nonetheless, this latter approach would overcome some major constraints such as the need for simultaneous acquisition in binocular stereo or handling dynamic objects in depth-from-motion approaches.
Although a geometrically ambiguous problem, Convolutional Neural Networks (CNNs) achieved outstanding results in monocular depth estimation by casting it as a learning task in both supervised and unsupervised manner.
In particular, the latter paradigm addresses the hunger for data typical of deep learning tasks by training networks to produce a depth representation minimizing the warping error between images acquired from multiple points of view rather than the error with respect to difficult to source ground-truth depth labels. 
In this field, the work of Godard et al. \cite{godard2017unsupervised} represents state-of-the-art for unsupervised monocular depth estimation. Deploying stereo imagery for training, a CNN learns to infer disparity from a single reference image and warps the target image accordingly to minimize the appearance error between the warped and the reference image.
This strategy yields state-of-the-art performance \cite{godard2017unsupervised,zhan2018unsupervised}. The CNN is trained to infer disparity from a single reference image and the target image is warped accordingly minimizing the appearance error between warped and reference image.
This way, the depth representation learned by the network is affected by artifacts in specific image regions inherited from the stereo setup (e.g., the left border using the left image as the reference) and in occluded areas.
The post-processing step proposed in \cite{godard2017unsupervised} partially compensates for these artifacts. However, it requires a double forward of the input image and its horizontally flipped version thus obtaining two predictions with artifacts, respectively, on the left and right side of depth discontinuities. Such issues are softened in the final map at the cost of doubling processing time and memory footprint.

In this paper, we propose to explicitly take into account these artifacts training our network on imagery acquired by a trinocular setup. By assuming the availability of three horizontally aligned images at training time, our network learns to process the frame in the middle and produce inverse depth (i.e., disparity) maps according to all the available viewpoints. By doing so, we can attenuate the aforementioned occlusion artifacts because they occur in different regions of the estimated outputs.
However, since trinocular setups are generally uncommon and hence datasets seldom available, we will show how to rely on popular stereo datasets such as CityScapes \cite{cordts2016cityscapes} and KITTI \cite{KITTI_RAW} to enforce our trinocular training assumption. Experimental results clearly prove that, deploying stereo pairs with a smart strategy aimed at emulating a trinocular setup, our  \underline{Three}-view \underline{Net}work (3Net) is able anyway to learn a three-view representation of the scene as shown intuitively in Figure \ref{fig:abstract} and how it leads to more robust monocular depth estimation compared to state-of-the-art methods trained on the same binocular stereo pair with a conventional paradigm.
Figure \ref{fig:abstract} highlights the behavior of 3Net: we can see how disparity maps (b) and (d), from the point of view of two frames respectively on the left and right side of the reference image, show mirrored artifacts in occluded regions. Combining the two opposite views enables to compensate for these issues and produces a more accurate map (c) centered on the reference frame. Please note that KITTI does not explicitly contain trinocular views as those shown in Figure \ref{fig:abstract} and that this behavior is learned by 3Net trained only on standard binocular data. Indeed, images and depth maps in (b) and (d) are inferred by our network.
Exhaustive experimental results on the KITTI 2015 stereo dataset \cite{KITTI_2015} and the Eigen split \cite{eigen2014depth} of the KITTI dataset \cite{KITTI_RAW} clearly show that 3Net, trained on standard binocular stereo pairs, improves state-of-the-art methods for unsupervised monocular depth estimation, regardless of the cues deployed for training.

\section{Related Work}
\label{Related_work}

In this section, we review the literature concerning single view depth estimation in both supervised and unsupervised manner. Moreover, we also consider early works on multi-baseline stereo setup being these approaches relevant to our proposal.

\textbf{Supervised depth-from-mono.} 
The following techniques share the need for difficult to source ground-truth depth measurements for training, thus posing a substantial limitation to their practical deployment. 
Saxena et al. \cite{saxena2009make3d} estimated depth and local planes using a MRF framework. Ladicky et al. \cite{ladicky2014pulling} proved that semantic can help depth estimation using a boosting classifier. More recently, CNN has emerged as mainstream strategy to estimate depth from a single image \cite{eigen2014depth,liu2016learning,laina2016deeper,li2015depth}.
Ummenhofer et al. \cite{ummenhofer2017demon} proposed DeMoN, a deep model to infer both depth and ego-motion from a pair of subsequent frames acquired by a single camera. Fu et al. \cite{fu2018supervised} introduced a novel strategy to discretize depth and cast the learning process as an ordinal regression problem, while Xu et al. \cite{xu2018supervised} integrated CRF models into deep architectures to improve depth prediction. Luo et al. \cite{luo2018supervised} formulated the monocular depth estimation problem as a view synthesis procedure followed by a deep stereo matching approach. Kumar et al. \cite{kumar2018recurrent} introduced a Recurrent Neural Network (RNN) aimed at predicting depth from monocular video sequences.  Lastly, Atapour et al. \cite{atapour2018supervised} exploited image style transfer and adversarial training to predict depth from real images training the network on a large amount of synthetic data.

\textbf{Unsupervised depth-from-mono.} 
Rethinking depth estimation as an image reconstruction task allowed to avoid the need for ground-truth depth labels and some works concerned with view synthesis paved the way for this purpose. Flynn et al. \cite{flynn2016deepstereo} proposed DeepStereo to generate new points of view training on images acquired by multiple cameras. Xie et al. \cite{xie2016deep3d} trained their Deep3D framework to create, from a single image, a target frame paired with the input according to a stereo setup by learning a disparity representation.

Unsupervised monocular depth estimation methods can be broadly categorized into two main categories according to the cues used to replace ground-truth labels. The first one \cite{garg2016unsupervised,godard2017unsupervised} leverages images with known relative camera pose, typically acquired by a calibrated stereo rig, following the strategy outlined by Deep3D \cite{xie2016deep3d}. A seminal work using this methodology was proposed by Garg et al. \cite{garg2016unsupervised}. Godard et al. \cite{godard2017unsupervised} deploying spatial transformer networks \cite{jaderberg2015spatial} and left-right consistency were able to notably improve depth accuracy. More compact models \cite{pydnet18} can be trained the same way and deployed on embedded systems as well. 

The second category concerns the use of imagery acquired by an unconstrained moving camera \cite{zhou2017unsupervised,mahjourian2018unsupervised}. Differently, from the previous methodology, temporally adjacent frames acquired by a single moving camera may contain dynamic objects that need to be explicitly handled during re-projection. Moreover, camera pose is unknown and needs to be estimated together with depth. On the other hand, such a strategy does not require a stereo camera to collect training samples. On this track, Zhou et al. \cite{zhou2017unsupervised} proposed a model to infer depth from unconstrained video sequences by computing a reconstruction loss between subsequent frames and predicting, at the same time, the relative pose between them. This strategy was improved by Mahjourian et al. \cite{mahjourian2018unsupervised} thanks to a 3D point-cloud alignment loss and by Wang et al. \cite{wang2018unsupervised} including a differentiable implementation of Direct Visual Odometry (DVO) with a novel depth normalization strategy. Yin et al. \cite{yin2018geonet} proposed GeoNet, a framework for depth and optical flow estimation from monocular sequences. 
Finally, we mention the work of Zhan et al. \cite{zhan2018unsupervised} which combined both strategies (i.e., training on stereo sequences) and the semi-supervised works of Kuznietsov et al. \cite{Kuznietsov_2017_CVPR} and Kumar et al. \cite{kumar2018gan}.

\textbf{Multi-baseline stereo}. It is generally recognized that using more than two views has the potential to improve the quality of depth estimation. An early work concerning multi-camera stereo was proposed by Minoru and Akira \cite{ito1986three} deploying a triangular rig, while Okutomi and Kanade \cite{okutomi1993multiple} achieved accurate depth measurements combining stereo from multiple baseline cameras horizontally aligned. Kang et al. \cite{kang2001handling} proposed a method to handle the increasing number of occlusions occurring in multi-view stereo setup, while Ayache and Lustman \cite{Ayache1991} designed a three cameras rig for robotic applications and Garcia et al. \cite{Garcia2002} proposed a pose detection algorithm based on a trinocular stereo system.
In the last decade, along with the availability of off-the-shelf stereo cameras  (e.g., Intel RealSense) some multi-baseline stereo systems too were commercially made available. For instance, the Bumblebee XB3 was used to acquire the RobotCar dataset \cite{RobotCarDatasetIJRR}, counting \emph{millions} of images acquired driving for about 1000 Km. Honneger et al. \cite{HoneggerICRA2017} developed a multi-baseline camera with on-board FPGA, enabling real-time processing of dense disparity maps. Therefore, a trinocular stereo configuration for training, like the one we advocate in our work, would be undoubtedly feasible. Nonetheless, our strategy is feasible and useful even with conventional binocular datasets.

\begin{figure}
    \centering
    \begin{tabular}{cc}
        \multicolumn{2}{c}{\includegraphics[width=0.48\textwidth]{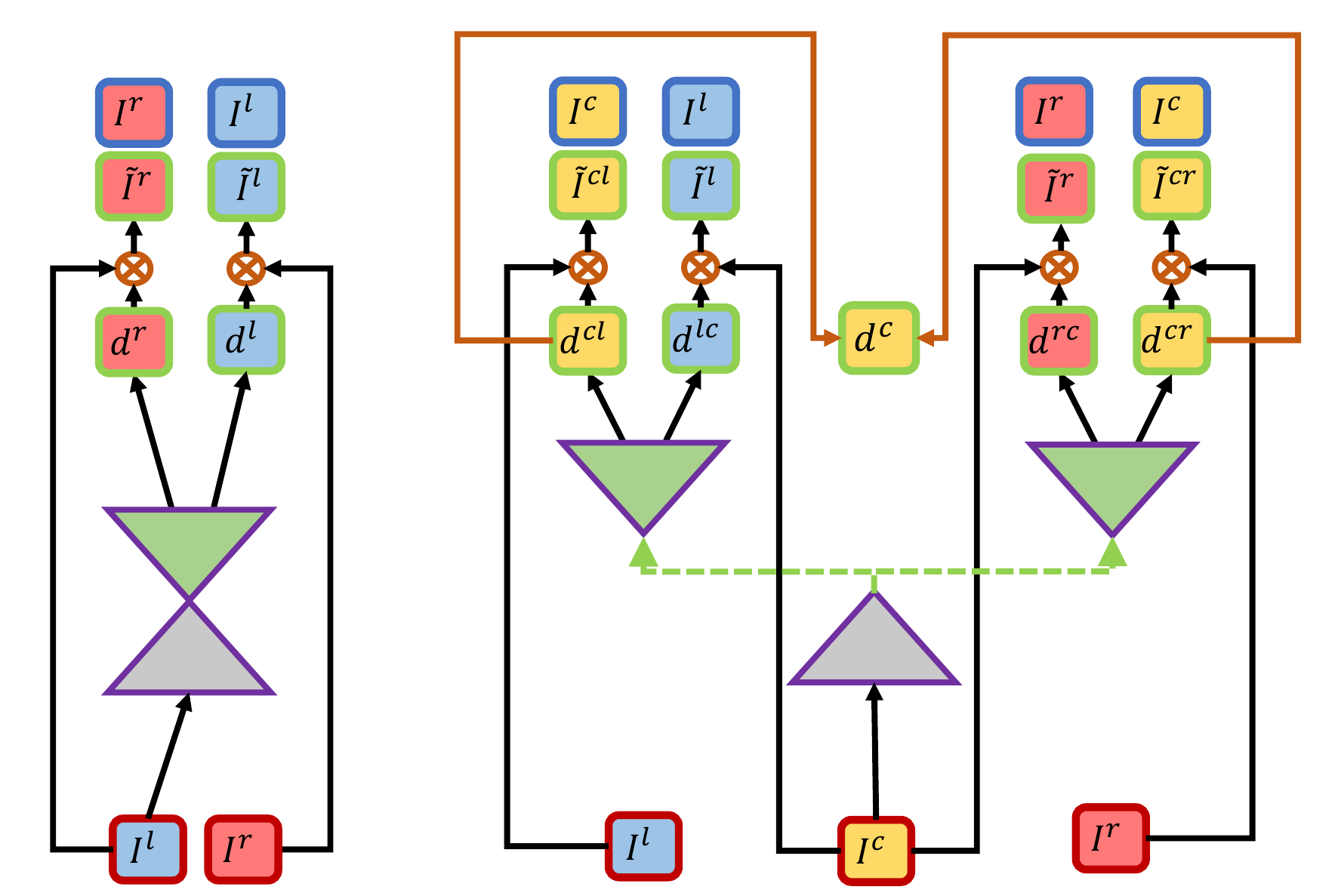} }\\
        \hspace{30pt} a) & \hspace{20pt} b)\\
    \end{tabular}
    \caption{Training frameworks enforcing a) binocular \cite{godard2017unsupervised} and b) trinocular assumptions.}
    \label{fig:trino}
\end{figure}

\section{Method overview}

In this section, we propose a framework aimed at enforcing a trinocular assumption for training in an unsupervised manner a network for monocular depth estimation. We will outline the rationale behind this choice and the differences with known techniques in the literature. Then, deploying a conventional binocular stereo dataset, we will show how our strategy allows advancing state-of-the-art.

\subsection{Trinocular assumption and network design}

While traditional depth-from-mono frameworks learn to estimate $d(I)$ from an input image $I$ by minimizing the prediction error with respect to a ground-truth map $\hat{d}(I)$ whose pixels are labelled with real depth measurements, the introduction of image-reconstruction based losses moved this task to an unsupervised learning paradigm. In particular, estimated depth is used to project across different points of view exploiting 3D geometry and camera pose thus obtaining supervision signals through the minimization of the re-projection error. 
According to the literature reviewed in Section \ref{Related_work}, the training methodology based on images acquired with a stereo camera, as in \cite{garg2016unsupervised,godard2017unsupervised}, removes the need to infer pose estimation required when gathering data with a single unconstrained camera.

\begin{figure}
    \centering
    \includegraphics[width=0.42\textwidth]{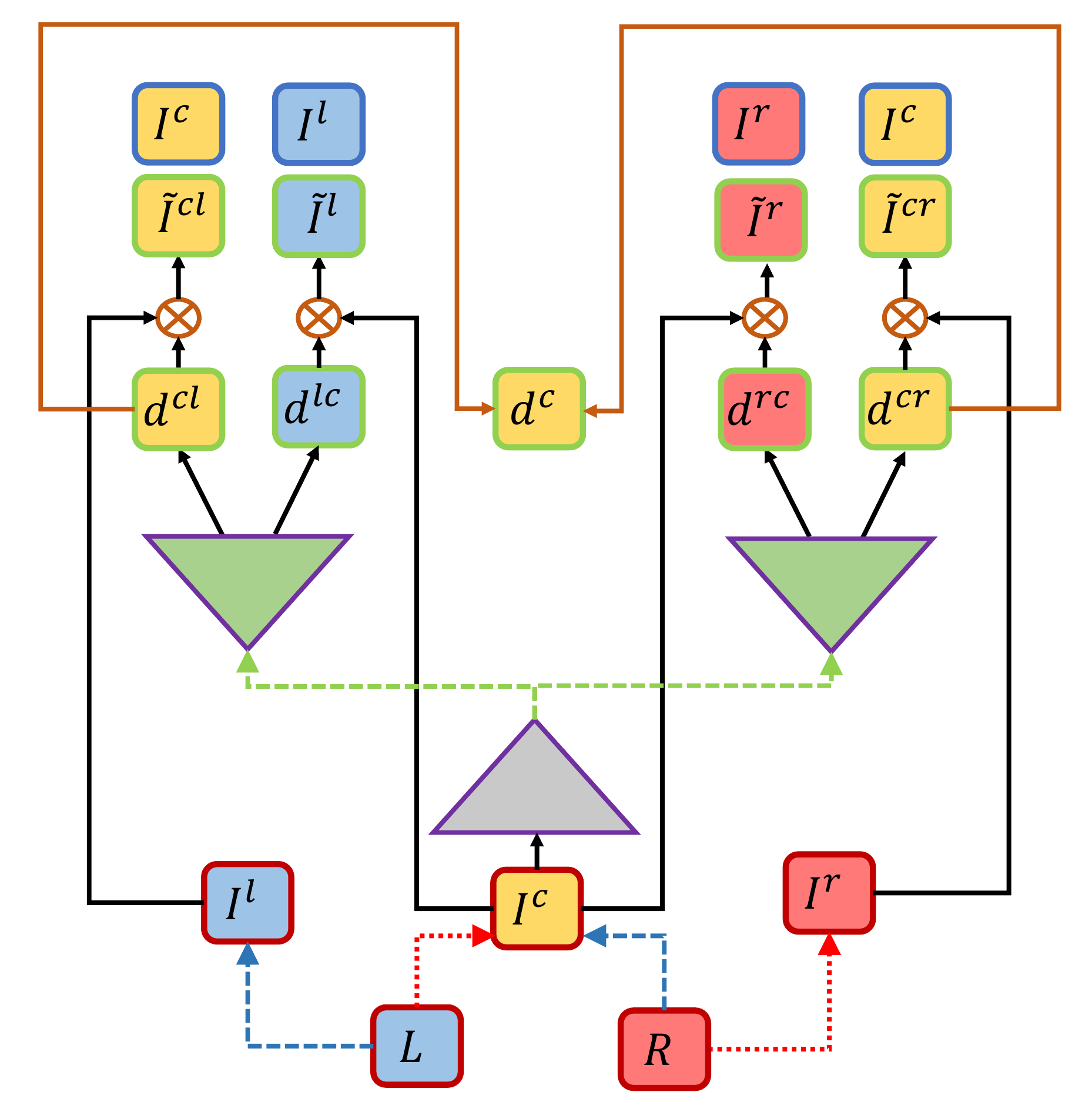}
    \caption{Scheme of interleaved training. A binocular stereo pair is used to train the network enforcing a trinocular assumption by first setting $L\rightarrow I^l$, $R\rightarrow I^c$ (blue arrows) and optimizing the model according to losses on $\tilde{I}^l$, $\tilde{I}^{cl}$, then setting $L\rightarrow I^c $, $R\rightarrow I^r$ (red arrows) and optimizing the model according to losses on $\tilde{I}^{cr}$, $\tilde{I}^r$. }
    \label{fig:fake-trino}
\end{figure}

Coaching a CNN to infer depth emulating a stereo system for training introduces artifacts in the learned representation (i.e., disparity) intrinsically because of well-known issues arising when dealing with pixels having no direct matches across the two images, such as on left border or occlusions. Godard et al. \cite{godard2017unsupervised} deal with this problem using a simple, yet effective, trick. By processing a horizontally flipped input image and then back-flipping the result, artifacts will appear on the opposite side w.r.t. the result obtained on the un-flipped frame (e.g., on the right border rather than on the left). Thus, combining the two predictions allows removing artifacts partially. Nevertheless, this strategy requires two forwards, hence doubling memory footprint and runtime, which would not be necessary if the CNN could learn to estimate disparity concerning a frame acquired on the left w.r.t reference image.
Guided by this intuition, we rethink the training protocol of \cite{godard2017unsupervised} to exploit a trinocular configuration, on which the image we want to learn the depth of is the central frame of three horizontally aligned points of view.
Figure \ref{fig:trino} gives an overview of our framework b) and the one by Godard et al. \cite{godard2017unsupervised} leveraging binocular stereo a). While a) trains the network to estimate a depth representation for $I^l$ by means of disparity map $d^l$, used to warp $I^r$ to $\tilde{I}^l$ and measure the appearance difference with $I^l$, we process $I^c$ to obtain $d^{lc}$ and  $d^{cr}$, disparity maps assuming as target $I^l$ and $I^r$, then we warp these latter two images to obtain $\tilde{I}^{cl}$ and $\tilde{I}^{cr}$ to finally compute supervision signals as re-projection error w.r.t. $I^c$. Finally, in our framework, $d^{lc}$ and $d^{cl}$ are combined to attenuate occlusions and obtain the final $d^c$ from a single forward pass, conversely to \cite{godard2017unsupervised} which requires two forwards.
Eventually, as \cite{godard2017unsupervised} estimates $d^r$ to enforce losses between $\tilde{I}^r$,$I^r$ and the LRC consistency, our network generates $d^{lc}$ and $d^{rc}$ to exploit losses between $\tilde{I}^{l}$,$I^l$ and $\tilde{I}^{r}$,$I^r$.

Figure \ref{fig:trino} also highlights a further main difference between the two frameworks. While a traditional UNet architecture is used by previous works in literature \cite{zhou2017unsupervised,godard2017unsupervised}, we build two separate decoders respectively in charge of estimating $d^{cl}$ and $d^{cr}$ separately. This strategy adds a negligible overhead regarding memory and runtime requirements, being the encoder the most computationally expensive module of the framework (i.e., the decoder mostly applies upsampling operations). According to our experiments, training a single decoder to infer a disparity representation for both points of view yields slightly worse results.

\subsection{Interleaved training for binocular images}
\label{sec:interleaved}

To effectively learn mirrored representation and compensate for occlusions/borders, the framework outlined so far relies on a set of three horizontally aligned images at training time. Although sensors designed to acquire such imagery are currently available, for instance the aforementioned Bumblebee XB3, it is still quite uncommon to find publicly available images obtained in such configuration. Indeed, in this sense, the Oxford RobotCar dataset \cite{RobotCarDatasetIJRR} represents an exception providing a large amount of street scenes captured with the trinocular XB3 sensor. Unfortunately, the provided calibration parameters only allow obtaining aligned views between left-right and center-right cameras, hence not permitting to align the three views as we desire.
Nonetheless, we describe in this section how to train our framework leveraging the proposed trinocular assumption with a much more common binocular setup (e.g., KITTI dataset).
Given a stereo pair made of images $L$ and $R$, Figure \ref{fig:fake-trino} depicts how to enforce the trinocular assumption by scheduling an \emph{interleaved training} of the network.
We update the parameters of the network by optimizing its four outputs $d^{cl}, d^{lc}, d^{rc}$ and $d^{cr}$ in two steps:

\begin{enumerate}

\item Firstly, we assign $L$ to $I^l$ and $R$ to $I^c$ as shown by the blue arrows in Figure \ref{fig:fake-trino}. In other words, we assume that the stereo pair represents the left and center images of a \emph{virtual} trinocular system in which the right frame is missing. In this case, we use as supervision signal the reconstruction error between $\tilde{I}^{cl},I^c$ and $\tilde{I}^l,I^l$, producing gradients that flow to the left decoder and the encoder.

\item Then, as shown by the red arrows in Figure \ref{fig:fake-trino} we change the role of $L$ and $R$ assuming them, respectively, as $I^c$ and $I^r$. In this phase, we suppose to have the center and right images available hence implicitly assuming that in our virtual trinocular system the left image is missing. Thus, using the supervision given by re-projection errors on pairs $\tilde{I}^r,I^r$ and $\tilde{I}^{cr},I^c$, we optimize the parameters of the right decoder and the (shared) encoder. 

\end{enumerate}

It is worth to note that, following this protocol, every time we run a training iteration on a stereo pair the network learns all the depth representations output of our framework. Moreover, the two learned disparity pairs from the two views are obtained according to the same baseline (i.e., the same of the training stereo pairs), making them consistent and hence easy to combine in $d^c$.
Therefore, the network learns a trinocular representation even if it actually never sees the scene with such setup. Indeed, this strategy is very effective as supported by experimental evidence in Section \ref{Experiments}.

\section{Implementation details}

In this section, we provide a detailed description of our framework, designed with the TensorFlow APIs. The source code is available at \url{https://github.com/mattpoggi/3net}.

\subsection{Network architecture}

For our 3Net we follow a quite established design strategy adopted by other methods in this field \cite{liu2016learning,laina2016deeper,zhou2017unsupervised,godard2017unsupervised}, based on an encoder-decoder architecture. The peculiarity of our approach consists in two different decoders, as depicted in Figure \ref{fig:fake-trino}, in charge of learning disparity representations w.r.t. two points of view located respectively on the left and right side of the input image. 
In our network, depicted in Figure \ref{fig:fake-trino}, each decoder generates outputs at four different scales, respectively: full, half, quarter and $\frac{1}{8}$ resolution.
As encoder, we tested VGG \cite{simonyan2014very} and ResNet50 \cite{he2016deep} to obtain the most fair and complete comparison w.r.t. \cite{godard2017unsupervised}, being it our baseline and state-of-the-art.
To obtain the final map $d_c$, we merge the contribution of $d_{cl}$ and $d_{cr}$ using the same post-processing procedure applied in \cite{godard2017unsupervised}, thus keeping 5\% left-most pixels from $d_{cl}$, 5\% right-most from $d_{cr}$ and averaging the remaining ones.

\subsection{Training losses}

We train 3Net to minimize a multi-component loss made of appearance, smoothness and consistency-check terms similarly to \cite{godard2017unsupervised}, namely $\mathcal{L}_{ap}, \mathcal{L}_{ds}$ and $\mathcal{L}_{lcr}$. 

\begin{equation}
\mathcal{L}_{total} = \beta_{ap}(\mathcal{L}_{ap}) + \beta_{ds}(\mathcal{L}_{ds}) + \beta_{lcr}(\mathcal{L}_{lcr})
\label{eq:total}
\end{equation}
The first term uses a weighted sum of SSIM \cite{SSIM} and L1 between all four warped pairs and real images as shown on top of Figure \ref{fig:fake-trino}. The second applies an edge aware smoothness constraint to estimated disparities $d^{cl}, d^{lc}, d^{rc}$ and $d^{cr}$ as described in \cite{godard2017unsupervised}.
Finally, the consistency-check term includes left-right losses between pairs $d^{cl}, d^{lc}$.

\begin{equation}
\mathcal{L}_{lcr} = \mathcal{L}_{lr}(d^{cl},d^{lc}) + \mathcal{L}_{lr}(d^{cr},d^{rc})
\label{eq:cc}
\end{equation}
For a detailed description of $\mathcal{L}_{ap}, \mathcal{L}_{ds}$ and $\mathcal{L}_{lr}$ please refer to \cite{godard2017unsupervised} or our supplementary material.

Thus, according to the interleaved training schedule described in Section \ref{sec:interleaved}, we optimize 3Net splitting the function \ref{eq:total} into two sub-losses $\mathcal{L}_{p_1},\mathcal{L}_{p_2}$ deployed in the two different phases:

\begin{align}
\begin{split}
\mathcal{L}_{p_1} = &\beta_{ap}(\mathcal{L}_{ap}(\tilde{I}^{cl},I^c) + \mathcal{L}_{ap}(\tilde{I}^l,I^l)) \\
&+ \beta_{ds}(\mathcal{L}_{ds}(d^{cl},I^c) + \mathcal{L}_{ds}(d^{lc},I^l)) \\
&+ \beta_{lcr}(\mathcal{L}_{lr}(d^{cl},d^{lc}) 
\end{split}
\label{eq:step1}
\end{align}

\begin{align}
\begin{split}
\mathcal{L}_{p_2} = &\beta_{ap}(\mathcal{L}_{ap}(\tilde{I}^{cr},I^c) + \mathcal{L}_{ap}(\tilde{I}^r,I^r)) \\
&+ \beta_{ds}(\mathcal{L}_{ds}(d^{cr},I^c) + \mathcal{L}_{ds}(d^{rc},I^l)) \\
&+ \beta_{lcr}(\mathcal{L}_{lr}(d^{cr},d^{lr}) 
\end{split}
\label{eq:step2}
\end{align}

We also evaluated an additional loss term $\mathcal{L}_{cc} = |d^{cl} - d^{cr}|$ to enforce consistency between depth representation centered on $I^c$, being the baseline equal on both directions. However, this term propagates occlusions artifacts between the two depth maps leading to worse results.
We point out that despite the interleaved training protocol outlined, in any phase the outcome of 3Net always consists of four depth maps $d^{cl}, d^{lc}, d^{rc}$ and $d^{cr}$. Of course, this happens at testing/inference time as well, when 3Net is fed with a single image. 
Considering that decoders outputs depth maps at four scales, all losses are computed for each of them as in \cite{godard2017unsupervised}.

\begin{table*}[t]
\centering
\scalebox{0.85}{
\begin{tabular}{c|c|ccccc|ccc|c}
\multicolumn{3}{c}{} & \multicolumn{2}{c}{\cellcolor{LightYellow} Proposed method} & \multicolumn{2}{c}{\cellcolor{blue!25} Lower is better}
 & \multicolumn{2}{c}{\cellcolor{LightCyan} Higher is better} & \multicolumn{1}{c}{} \\
\hline
Method & Train set & \cellcolor{blue!25} Abs Rel & \cellcolor{blue!25} Sq Rel & \cellcolor{blue!25} RMSE & \cellcolor{blue!25} RMSE log & \cellcolor{blue!25} D1-all &  \cellcolor{LightCyan}$\delta<$1.25 &  \cellcolor{LightCyan}$\delta<1.25^2$ & \cellcolor{LightCyan}$\delta<1.25^3$ & Forwards\\
\hline
Godard et al. \cite{godard2017unsupervised} & K & 0.124 & 1.388 & 6.125 & 0.217 & 30.272 & 0.841 & 0.936 & 0.975 & $\times$1 \\
\cellcolor{LightYellow}3Net  & K & 0.119 & 1.201 & 5.888 & 0.208 & 31.851 & 0.844 & 0.941 & 0.978 & $\times$1\\
Godard et al. \cite{godard2017unsupervised} + pp & K & 0.117 & 1.177 & 5.804 & 0.206 & 29.945 & 0.848 & 0.943 & 0.977 & $\times$2\\
\cellcolor{LightYellow}3Net + pp  & K & 0.114 & 1.088 & 5.756 & 0.203 & 31.141 & 0.848 & 0.944 & 0.979 & $\times$2\\
Godard et al. \cite{godard2017unsupervised} & CS+K & 0.104 & 1.070 & 5.417 & 0.188 & 25.523 & 0.875 & 0.956 & 0.983 & $\times$1\\
\cellcolor{LightYellow}3Net  & CS+K & 0.101 & 0.954 & 5.211 & 0.181 & 24.632 & 0.875 & 0.958 & 0.985 & $\times$1\\
Godard et al. \cite{godard2017unsupervised} + pp & CS+K & 0.100 & 0.934 & 5.141 & 0.178 & 25.077 & 0.878 & \textbf{0.961} & \textbf{0.986} & $\times$2\\
\cellcolor{LightYellow}3Net + pp  & CS+K & \textbf{0.097} & \textbf{0.893} & \textbf{5.079} & \textbf{0.176} & \textbf{23.867} & \textbf{0.881} & \textbf{0.961} & \textbf{0.986} & $\times$2\\
\hline
\end{tabular}
}
\caption{Comparison between 3Net and \cite{godard2017unsupervised}, both using VGG as encoder, on KITTI 2015 training dataset \cite{KITTI_2015}.}
\label{table:kitti}
\end{table*}

\subsection{Training protocol}

We assume as baseline the framework proposed by Godard et al. \cite{godard2017unsupervised} using a binocular setup for unsupervised training. For a fair comparison, we train our models following the same guidelines reported in \cite{godard2017unsupervised}. In particular, we use CityScapes \cite{cordts2016cityscapes} (CS) and KITTI raw sequences \cite{KITTI_RAW} datasets for training, this latter sub-sampled according to two training splits of data \cite{godard2017unsupervised,eigen2014depth} to be able to compare our results with any recent works in this field using unsupervised learning. We refer to these two subsets as KITTI split (K) and Eigen split (E) \cite{eigen2014depth}. The three training sets count respectively about 23k, 29k and 22.6k stereo pairs. As pointed out by first works using image reconstruction losses \cite{godard2017unsupervised,zhou2017unsupervised}, training on different datasets helps the network to achieve higher-quality results. Therefore, to better assess the performance of each method, we report experimental results training the networks on K or E. Moreover, we also report results training on CityScapes and then fine tuning on K or E (respectively, referred to as CS+K and CS+E in the tables).
Consistently with \cite{godard2017unsupervised}, we run 50 epochs of training on each single dataset using a batch size of 8 and input resized to $256\times512$. We use Adam optimizer \cite{kingma2014adam} with $\beta_1 = 0.9, \beta_2 = 0.999$ and $\varepsilon = 10^{-8}$, setting an initial learning rate of $10^{-4}$ halved after 30 and 40 epochs. We maintain the same hyperparameters configuration for $\beta_{ap}, \beta_{ds}$ and $\beta_{lrc}$ defined in \cite{godard2017unsupervised} and the same data augmentation procedure as well.

\section{Experimental results}
\label{Experiments}

In this section, we assess the performance of our 3Net framework with respect to state-of-the-art. In all our tests, we report 7 main metrics measuring the average depth error (Abs Rel, Sq Rel, RMSE and RMSE log, the lower the better) and three accuracy scores ($\delta < 1.25, \delta < 1.25^2$ and $\delta < 1.25^3$, the higher the better), 
First, we report experiments on the K split assuming Godard et al. \cite{godard2017unsupervised} as baseline. 
Then, we exhaustively compare 3Net with top performing unsupervised frameworks for depth-from-mono estimation, highlighting how our proposal is state-of-the-art. It is worth stressing that the proposed interleaved training procedure of 3Net, described in Section \ref{sec:interleaved}, allows for a fair comparison with any other method included in our evaluation being all trained exactly on the same (binocular) datasets. Finally, we report qualitative results concerning the trinocular representation learned by 3Net.

\begin{figure}
    \centering
    \begin{overpic}[width=0.42\textwidth]{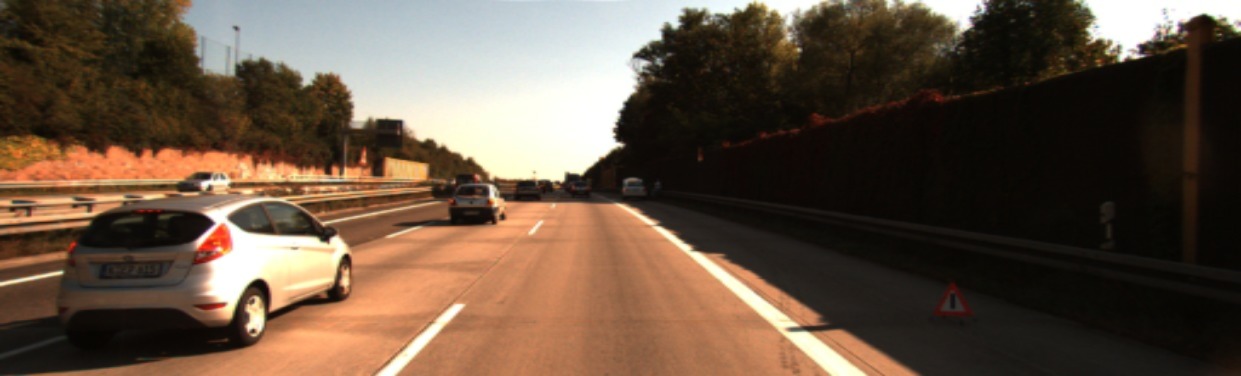}
    \put (92,4) {$\displaystyle\textcolor{white}{\textbf{(a)}}$}
    \end{overpic} \\
    \begin{overpic}[width=0.42\textwidth]{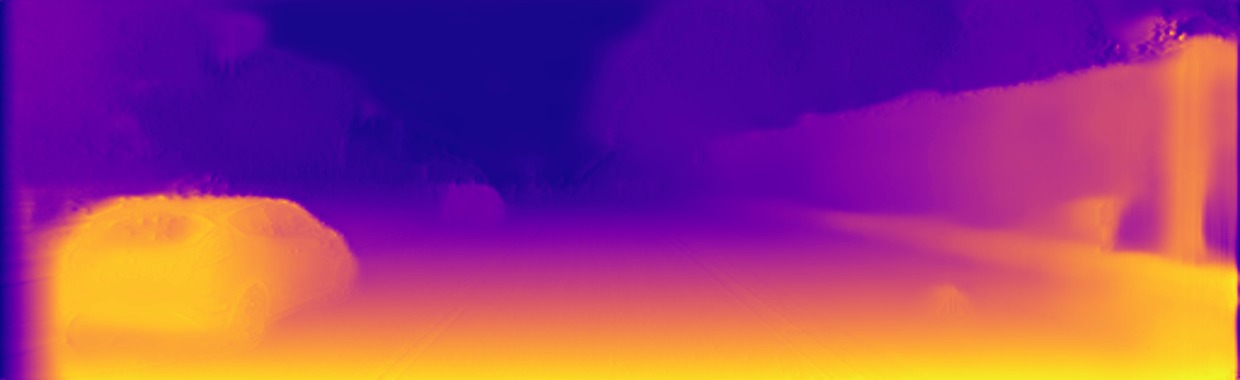}
    \put (92,4) {$\displaystyle\textcolor{white}{\textbf{(b)}}$}
    \end{overpic} \\
    \begin{overpic}[width=0.42\textwidth]{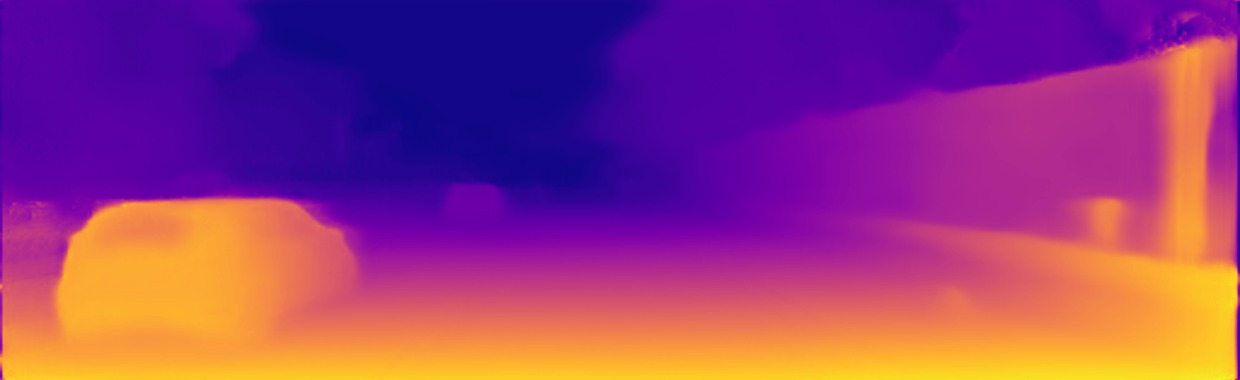}
    \put (92,4) {$\displaystyle\textcolor{white}{\textbf{(c)}}$}
    \end{overpic} \\
    \caption{Depth maps predicted from input image (a) by Godard et al. \cite{godard2017unsupervised} (b) and 3Net (c) running a single forward pass.}
    \label{fig:godard-vs-3net}
\end{figure}

\begin{table*}[t]
\centering
\scalebox{0.80}{
\begin{tabular}{c|c|c|cccc|ccc}
\multicolumn{3}{c}{} & \multicolumn{2}{c}{\cellcolor{LightYellow} Proposed method} &\multicolumn{2}{c}{\cellcolor{blue!25} Lower is better}
 & \multicolumn{2}{c}{\cellcolor{LightCyan} Higher is better} \\
\hline
Method & Supervision & Train set & \cellcolor{blue!25} Abs Rel & \cellcolor{blue!25} Sq Rel & \cellcolor{blue!25} RMSE & \cellcolor{blue!25} RMSE log &  \cellcolor{LightCyan}$\delta<$1.25 &  \cellcolor{LightCyan}$\delta<1.25^2$ & \cellcolor{LightCyan}$\delta<1.25^3$ \\
\hline
Kumar et al. \cite{kumar2018gan} (photo. + adv.) & Temporal & E& 0.211 & 1.980 & 6.154 & 0.264 & 0.732 & 0.898 & 0.959 \\
Zhou et al. \cite{zhou2017unsupervised} & Temporal & E& 0.208 & 1.768 & 6.856 & 0.283 & 0.678 & 0.885 & 0.957 \\
Zhou et al. \cite{zhou2017unsupervised} updated \cite{yin2018geonet} & Temporal & E& 0.183 & 1.595 & 6.709 & 0.270 & 0.734 & 0.902 & 0.959 \\
Mahjourian et al. \cite{mahjourian2018unsupervised} & Temporal & E & 0.163 & 1.240 & 6.220 & 0.250 & 0.762 & 0.916 & 0.968 \\
Yin et al. \cite{yin2018geonet} GeoNet & Temporal & E & 0.164 & 1.303 & 6.090 & 0.247 & 0.765 & 0.919 & 0.968 \\
Yin et al. \cite{yin2018geonet} GeoNet ResNet50 & Temporal & E & 0.155 & 1.296 & 5.857 & 0.233 & 0.793 & 0.931 & 0.973 \\
Wang et al. \cite{wang2018unsupervised} & Temporal & E & 0.151 & 1.257 & 5.583 & 0.228 & 0.810 & 0.936 & \textbf{0.974} \\
Poggi et al. \cite{pydnet18} PyD-Net (200) & Stereo & E & 0.153 & 1.363 & 6.030 & 0.252 & 0.789 & 0.918 & 0.963 \\
Godard et al. \cite{godard2017unsupervised} & Stereo & E & 0.148 & 1.344 & 5.927 & 0.247 & 0.803 & 0.922 & 0.964 \\
Zhan et al. \cite{zhan2018unsupervised} & Stereo+Temp. & E & 0.144 & 1.391 & 5.869 & 0.241 & 0.803 & 0.928 & 0.969 \\
\cellcolor{LightYellow}3Net  & Stereo & E & 0.142 & 1.207 & 5.702 & 0.240 & 0.809 & 0.928 & 0.967\\

Godard et al. \cite{godard2017unsupervised} ResNet50 & Stereo & E & 0.133 & 1.142 & 5.533 & 0.230 & 0.830 & 0.936 & 0.970 \\
\cellcolor{LightYellow}3Net ResNet50  & Stereo & E & 0.129 & 0.996 & 5.281 & 0.223 & 0.831 & 0.939 & \textbf{0.974} \\
Godard et al. \cite{godard2017unsupervised} ResNet50 + pp & Stereo & E & 0.128 & 1.038 & 5.355 & 0.223 & 0.833 & 0.939 & 0.972 \\
\cellcolor{LightYellow}3Net ResNet50 + pp  & Stereo & E & \textbf{0.126} & \textbf{0.961} & \textbf{5.205} & \textbf{0.220} & \textbf{0.835} & \textbf{0.941} & \textbf{0.974} \\

\hline
Zhou et al. \cite{zhou2017unsupervised} & Temporal & CS+E & 0.198 & 1.836 & 6.565 & 0.275 & 0.718 & 0.901 & 0.960 \\
Mahjourian et al. \cite{mahjourian2018unsupervised} & Temporal & CS+E & 0.159 & 1.231 & 5.912 & 0.243 & 0.784 & 0.923 &  0.970 \\
Yin et al. \cite{yin2018geonet} GeoNet ResNet50 & Temporal & CS+E & 0.153 & 1.328 & 5.737 & 0.232 & 0.802 & 0.934 & 0.972\\
Wang et al. \cite{wang2018unsupervised} & Temporal & CS+E & 0.148 & 1.187 & 5.496 & 0.226 & 0.812 & 0.938 & 0.975\\
Poggi et al. \cite{pydnet18} PyD-Net (200) & Stereo & CS+E & 0.146 & 1.291 & 5.907 & 0.245 & 0.801 & 0.926 & 0.967 \\
Godard et al. \cite{godard2017unsupervised} & Stereo & CS+E & 0.124 & 1.076 & 5.311 & 0.219 & 0.847 & 0.942 & 0.973 \\
\cellcolor{LightYellow}3Net  & Stereo & CS+E & 0.117 & 0.905 & 4.982 & 0.210 & 0.856 & 0.948 & 0.976 \\
Godard et al. \cite{godard2017unsupervised} ResNet50 & Stereo & CS+E & 0.121 & 1.037 &  5.212 & 0.216 & 0.854 & 0.944 & 0.973 \\
\cellcolor{LightYellow}3Net ResNet50  & Stereo & CS+E & 0.113 & 0.885 & 4.898 & 0.204 & 0.862 &  0.950 & 0.977 \\
Godard et al. \cite{godard2017unsupervised} ResNet50 + pp & Stereo & CS+E & 0.114 & 0.898 & 4.935 & 0.206 & 0.861 & 0.949 & 0.976 \\
\cellcolor{LightYellow}3Net ResNet50 + pp  & Stereo & CS+E & \textbf{0.111} & \textbf{0.849} & \textbf{4.822} &  \textbf{0.202} & \textbf{0.865} & \textbf{0.952} & \textbf{0.978} \\
\hline
\end{tabular}
}
\caption{Evaluation on the KITTI dataset \cite{KITTI_RAW} using the split of Eigen et al. \cite{eigen2014depth}, with maximum depth set to 80m. Results concerned with state-of-the-art techniques for unsupervised monocular depth estimation leveraging video sequences (Temporal), binocular stereo pairs (Stereo) and both cues (Stereo+Temp.).}
\label{table:eigen}
\end{table*}

\subsection{KITTI split} 

Table \ref{table:kitti} reports experimental results on the KITTI 2015 stereo dataset \cite{KITTI_2015}. The evaluation was carried out on 200 stereo pairs with available high quality ground-truth disparity annotations. 
Additionally, being the outputs of \cite{godard2017unsupervised} and 3Net disparity maps, in our evaluation we include the D1-all score representing the percentage of pixels having a disparity error larger than 3.

We compare the raw output $d_c$ of our network with the map predicted by Godard et al. with and without post-processing \cite{godard2017unsupervised} (namely ``+pp'' in the table) running, respectively, a single or two forwards of the network.
Moreover, since 3Net can benefit from the same refinement technique by running two predictions on $I^c$ and its horizontally flipped version, we also provide results for our network applying the same post-processing. Therefore, we estimate post-processed $d^{cl}$ and $d^{cr}$ before combining them into $d^c$. Anyway, we report for clarity in the last column of the table, the number of forwards required by each entry.

As reported on the first two rows of Table \ref{table:kitti}, training the networks on KITTI data only, our method outperforms the competitor on all metrics except D1-all when running a single forward and it performs very similar to the post-processed version of \cite{godard2017unsupervised} reported in the third row of the table. Rows 3 and 4 highlight that, performing two forwards and post-processing, 3Net + pp outperforms Godard et al. + pp again on all metrics except D1-all.

Previous works in literature \cite{zhou2017unsupervised,godard2017unsupervised,zhan2018unsupervised,wang2018unsupervised,yin2018geonet,mahjourian2018unsupervised} proved that transfer learning from CityScape dataset \cite{cordts2016cityscapes} to KITTI is beneficial and leads to more accurate depth estimation, thus we follow this guideline training on CS+K as well.
The last four rows of Table \ref{table:kitti} compare both frameworks with and without post-processing. Without post-processing, we can notice how pre-training on CityScapes dataset allows 3Net to outperform \cite{godard2017unsupervised} on all metrics including D1-all. In the last two rows, applying post-processing to the output of both models, 3Net outperforms the competitor on all metrics tying on $\delta<1.25^2$ and $\delta<1.25^3$.
Figure \ref{fig:godard-vs-3net} qualitatively shows depth maps predicted by \cite{godard2017unsupervised} (b) and 3Net (c) without applying any post-processing to better perceive the improvements lead by our framework.

Summarizing, experiments on the KITTI split highlighted how enforcing the trinocular assumption is more effective than leveraging a conventional stereo paradigm for training. Moreover, these results prove that stereo pairs can be used in a smarter way following our interleaving strategy.

\subsection{Eigen split} 

Table \ref{table:eigen} reports evaluation with the split of data of Eigen et al. \cite{eigen2014depth}, made of 697 images and relative depth measurements acquired with a Velodyne sensor. The table collects results concerning most recent works addressing unsupervised monocular depth estimation. For each method, we indicate the kind of supervision it leverages on: monocular sequences (\emph{Temporal}), stereo pairs (\emph{Stereo}) or stereo sequences (\emph{Stereo+Temp.}). We report results either training on E only or on CS+E, allowing to compare our scores with state-of-the-art approaches. We point out that all methods, including our proposal, are trained exactly on the same images and all of them \emph{see} the same scenes\footnote{Zhan et al. \cite{zhan2018unsupervised} report scores training on E only or after pre-training on NYU dataset \cite{SilbermanECCV12}. For fairness we report the first setup only.}.
On top, we report results for models trained on the Eigen split of data. For GeoNet \cite{yin2018geonet}, Godard et al. \cite{godard2017unsupervised} and our method we report results for both VGG and ResNet50 encoders.
We can notice that, in general, methods trained using stereo data usually outperform those trained on monocular video sequences, as evident from recent literature \cite{zhou2017unsupervised,godard2017unsupervised,zhan2018unsupervised,wang2018unsupervised,yin2018geonet,mahjourian2018unsupervised}. Zhan et al. \cite{zhan2018unsupervised} leveraging temporally adjacent stereo frames outperform, on most metrics, \cite{godard2017unsupervised}. Nevertheless, 3Net achieves better scores except for $\delta<1.25^3$ still without exploiting temporal supervision. 
This proves that a smarter deployment of binocular training samples, i.e. by applying our interleaved training to fulfill trinocular hypothesis, is an effective alternative to sequence supervision. It is worth to note that Wang et al. \cite{wang2018unsupervised} obtain better scores on most metrics (RMSE, RMSE log and $\delta$ metrics) w.r.t. \cite{godard2017unsupervised} and 3Net with the VGG encoder.
However, by switching to the ResNet50 encoder, Godard et al. \cite{godard2017unsupervised} overtakes most recent works that use \emph{Time} supervision \cite{wang2018unsupervised,yin2018geonet} with and without post-processing. Systematically, 3Net always outmatches \cite{godard2017unsupervised} and consequently all its competitors. In particular, we point out that 3Net ResNet50 without post-processing already achieves some better scores compared to Godard et al. ResNet50 + pp performing, respectively, a single and a double forward.

\begin{figure*}
    \centering
    \begin{overpic}[width=0.42\textwidth]{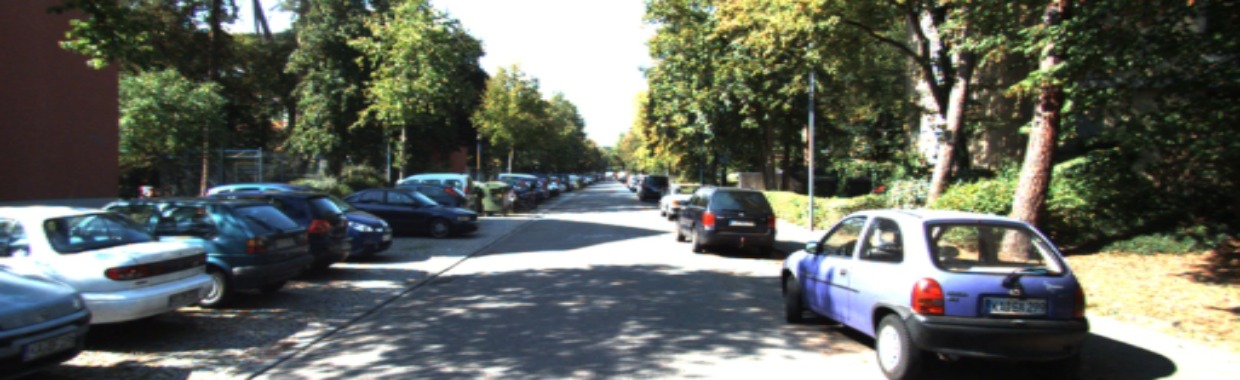}
    \put (2,4) {$\displaystyle\textcolor{white}{\textbf{(a)}}$}
    \end{overpic} 
    \begin{overpic}[width=0.42\textwidth]{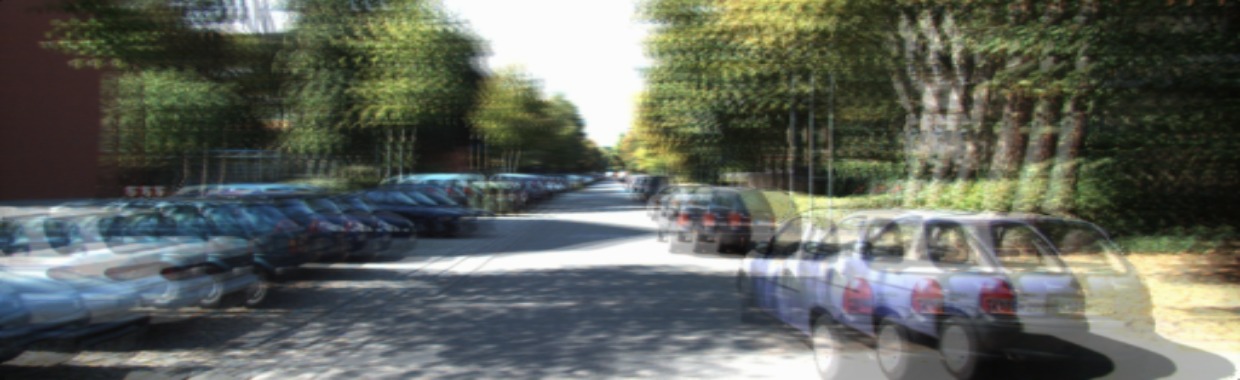}
    \put (2,4) {$\displaystyle\textcolor{white}{\textbf{(b)}}$}
    \end{overpic} \\
    \begin{overpic}[width=0.42\textwidth]{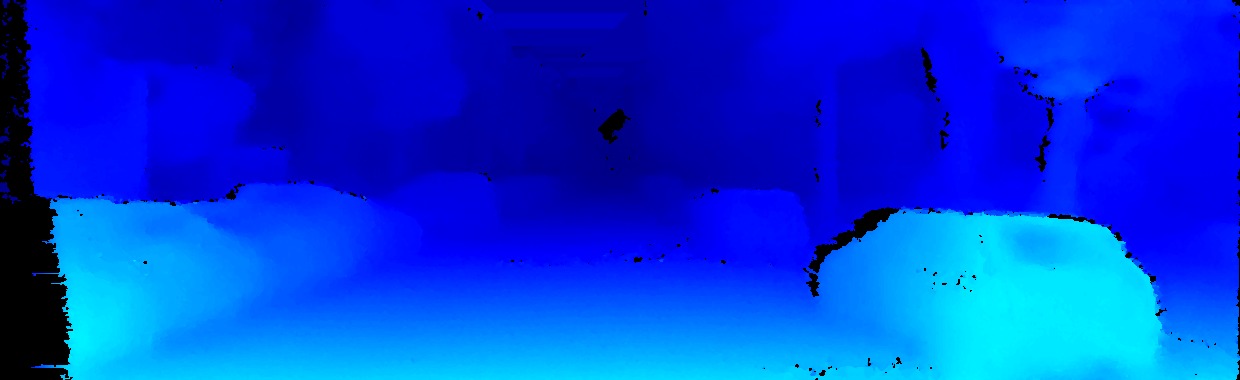}
    \put (2,4) {$\displaystyle\textcolor{white}{\textbf{(c)}}$}
    \end{overpic} 
    \begin{overpic}[width=0.42\textwidth]{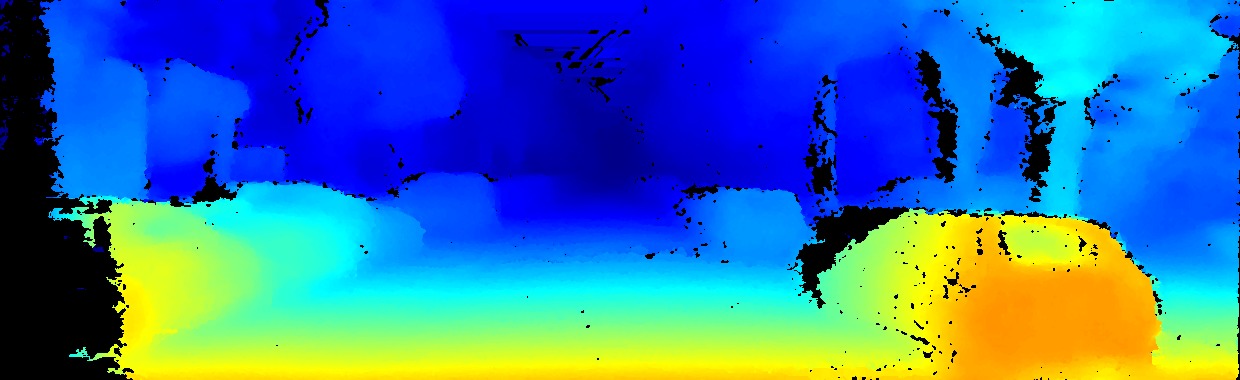}
    \put (2,4) {$\displaystyle\textcolor{white}{\textbf{(d)}}$}
    \end{overpic} \\
    \caption{Qualitative example of learned trinocular setup. Given a single, input image (a), 3Net can generate two additional points of view, shown superimposed to the real frame in (b). Running traditional stereo algorithms \cite{hirschmuller2005accurate}, assuming the left-most generated frame as reference, allows to obtain disparity maps with narrow (c) and wide (d) baseline (encoded with colormap jet). We point out that the center frame (a) is the only real image. More qualitative examples in the supplementary material.}
    \label{fig:multibaseline}
\end{figure*}

On the bottom of Table \ref{table:eigen}, we resume results achieved by models trained on CS+E. We observe the same trend highlighted in the previous experiments, being \cite{godard2017unsupervised} and our proposal the most effective solutions for this task thanks to stereo supervision.
In equal conditions, i.e. same encoder and number of forwards, 3Net always outperforms the framework of Godard et al. exploiting the trinocular assumption. Moreover, the proposed technique leads to major improvements such that 3Net VGG outperforms ResNet50 model by Godard et al. \cite{godard2017unsupervised} (rows 20$^{th}$ and 21$^{st}$), 3Net ResNet50 without post-processing achieves more accurate results than the best configuration ResNet50 + pp \cite{godard2017unsupervised} (rows 22$^{nd}$ and 23$^{rd}$) and finally 3Net ResNet50 + pp outmatches all known frameworks for unsupervised depth-from-mono estimation. These facts clearly highlight that our proposal is state-of-the-art. 

It is important to underline that the availability of a \emph{real} trinocular rig would most probably allow training a more accurate model, given the larger amount of images w.r.t. a binocular stereo rig. The interleaved training proposed in this paper allows to overcome the lack of trinocular training samples using binocular pairs and also allows for a more fair comparison with other techniques leveraging this latter configuration only. This fact proves that the effectiveness of our strategy is due to the rationale behind it and not driven by a more extensive availability of data.

\section{View synthesis}

Finally, we show through qualitative images some outcomes of 3Net obtainable exploiting the embodied trinocular assumption.\footnote{A video is available at \url{http://youtu.be/uMA5YWJME4M}}. 
A peculiar experiment allowed by our framework consists of generating three horizontally aligned views from a single input image. This feature is possible thanks to estimated $d^{lc}$ and $d^{rc}$, used to warp the input image towards two new viewpoints, respectively, on the left and the right.
In other words, given $I^c$ at test time we compute $\tilde{I}^l$ and $\tilde{I}^r$, producing a trinocular setup of horizontally aligned images.
Figure \ref{fig:multibaseline} shows an example of a single frame (a) taken from the KITTI dataset and how our network generates the three views superimposed in (b).
These views effectively enable to realize a multi-baseline setup. Thus we can run any stereo algorithm between the possible pairs. For instance, we run the Semi-Global Matching algorithm (SGM) \cite{hirschmuller2005accurate} between $\tilde{I}^l$ and $I^c$, showing the results in Figure \ref{fig:multibaseline} (c), then we run SGM between $\tilde{I}^l$ and $\tilde{I}^r$ obtaining the disparity map shown in (d).
The two disparity maps assume the same frame as the reference image ($\tilde{I}^l$) and two different targets, according to two different \emph{narrow} and \emph{wide} virtual baseline. The shorter baseline is learned from the KITTI acquisition rig while the longer one is inherited by our trinocular assumption although actually, it does not exist at all in the training set. This fact can be perceived by looking at the different disparity ranges encoded, with colormap jet, on (c) and (d). This feature of our network paves the way to exciting novel developments. For instance, a conceivable application would consist in the synthesis of \emph{augmented} stereo pairs to train CNNs for disparity inference \cite{mayer2016large,godard2017unsupervised} or to improve recent techniques such as single view stereo \cite{luo2018supervised}.

\section{Conclusions}

In this paper, we proposed a novel methodology for unsupervised training of a depth-from-mono CNN. By enforcing a trinocular assumption, we overcome some limitations due to binocular stereo images used as supervision and obtain a more accurate depth estimation with our 3Net architecture. Although three horizontally aligned views are seldom available, we proposed an interleaved training protocol allowing to leverage on traditional binocular datasets. This latter technique also ensures for a fair comparison w.r.t. all previous works and allows us to prove that 3Net outperforms all unsupervised techniques known in the literature, establishing itself as state-of-the-art.
Moreover, 3Net learns a trinocular representation of the world, making it suitable for image synthesis purposes and other interesting future developments. 

\section*{Acknowledgements} 

We gratefully acknowledge the support of NVIDIA Corporation with the donation of the Titan X GPU used for this research.

{\small
\bibliographystyle{ieee}
\bibliography{egbib}
}

\clearpage
	
	\section*{Supplementary material} 
	
	This document provides additional details and experimental results concerned with paper "Learning monocular depth estimation under unsupervised trinocular assumption". The supplementary material is organized as follows: Section \ref{sec:losses} reports detailed explanation of the loss functions used at training time, Section \ref{sec:pp} describes how we obtain $d^c$ with 3Net and how we post-process it, Section \ref{sec:50m} comments additional experiments on the Eigen split \cite{eigen2014depth} assuming as maximum depth 50 meters and Section \ref{sec:qualitative} collects additional qualitative results, Finally, Section \ref{sec:time} reports run time analysis for 3Net and \cite{godard2017unsupervised}.

	\section{Training losses}
	\label{sec:losses}
	
	In the paper, all loss functions are computed at four scales, ranging from full image resolution to $\frac{1}{8}$. The global loss function is defined as:
	
	\begin{equation}
	\mathcal{L}_{total} = \beta_{ap}(\mathcal{L}_{ap}) + \beta_{ds}(\mathcal{L}_{ds}) + \beta_{lcr}(\mathcal{L}_{lcr})
	\label{eq:total}
	\end{equation}
	
	where $\mathcal{L}_{ap}$, $\mathcal{L}_{ds}$ and $\mathcal{L}_{lcr}$ represent, respectively, the appearance, smoothness and consistency terms, while $\beta_{ap}$, $\beta_{ds}$ and $\beta_{lcr}$ are hyper-parameters. In particular, we set $\beta_{ap}=\beta_{lcr}=1$ and $\beta_{ds}=0.1$. 
	\newline
	
	\textbf{Appearance Loss.}
	It measures the reconstruction error between a warped image and the original one. It is obtained by a weighted sum of a SSIM based score \cite{SSIM} and a L1 distance over pixel intensities.
	
	\begin{align}
	\begin{split}
	\mathcal{L}_{ap}(I^l,I^r) = \frac{1}{N} \sum_{ij} & \alpha \frac{1 - SSIM(I_{ij}^{l}, \tilde{I_{ij}^{r}})}{2}\\ & + (1 - \alpha) || I_{ij}^{l} - \tilde{I_{ij}^{r}} ||
	\end{split}
	\label{eq:step2}
	\end{align}

	\begin{table*}[!htbp]
		\centering
		\scalebox{0.80}{
			\begin{tabular}{c|c|c|cccc|ccc}
				\multicolumn{3}{c}{} & \multicolumn{2}{c}{\cellcolor{LightYellow} Proposed method} &\multicolumn{2}{c}{\cellcolor{blue!25} Lower is better}
				& \multicolumn{2}{c}{\cellcolor{LightCyan} Higher is better} \\
				\hline
				Method & Supervision & Train set & \cellcolor{blue!25} Abs Rel & \cellcolor{blue!25} Sq Rel & \cellcolor{blue!25} RMSE & \cellcolor{blue!25} RMSE log &  \cellcolor{LightCyan}$\delta<$1.25 &  \cellcolor{LightCyan}$\delta<1.25^2$ & \cellcolor{LightCyan}$\delta<1.25^3$ \\
				\hline
				Zhou et al. \cite{zhou2017unsupervised} & Temporal & E& 0.201 & 1.391 & 5.181 & 0.264 & 0.696 & 0.900 & 0.966 \\
				
				Mahjourian et al. \cite{mahjourian2018unsupervised} & Temporal & E & 0.155 & 0.927 & 4.549 & 0.231 & 0.781 & 0.931 & 0.975 \\
				
				Zhan et al. \cite{zhan2018unsupervised} & Stereo+Temp. & E & 0.135 & 0.905 & 4.366 & 0.225 & 0.818 & 0.937 & 0.973 \\
				
				Godard et al. \cite{godard2017unsupervised} ResNet50 + pp & Stereo & E & 0.1217 & 0.7630 & 4.047 & 0.210 & 0.847 & 0.946 & 0.976 \\
				
				\cellcolor{LightYellow} 3Net ResNet50 + pp \textbf{(ours)} & Stereo & E & \textbf{0.1207} & \textbf{0.7185} & \textbf{3.968} & \textbf{0.208} & \textbf{0.849} & \textbf{0.948} & \textbf{0.977} \\
				
				\hline
				Zhou et al. \cite{zhou2017unsupervised} & Temporal & CS+E & 0.190 & 1.436 & 4.975 & 0.258 & 0.735 & 0.915 & 0.968 \\
				
				Mahjourian et al. \cite{mahjourian2018unsupervised} & Temporal & CS+E & 0.151 & 0.949 & 4.383 & 0.227 & 0.802 & 0.935 & 0.974 \\
				
				Poggi et al. \cite{pydnet18} PyD-Net (200) & Stereo & CS+E & 0.138 & 0.937 & 4.488 & 0.230 & 0.815 & 0.934 & 0.972 \\
				
				Godard et al. \cite{godard2017unsupervised} ResNet50 + pp & Stereo & CS+E & 0.108 & 0.657 & 3.729 & 0.194 & 0.873 & 0.954 & \textbf{0.979} \\
				
				\cellcolor{LightYellow} 3Net ResNet50 + pp \textbf{(ours)} & Stereo & CS+E & \textbf{0.091} & \textbf{0.572} & \textbf{3.459} &  \textbf{0.183} & \textbf{0.889} & \textbf{0.955} & \textbf{0.979} \\
				\hline
			\end{tabular}
		}
		\caption{Evaluation on the KITTI dataset \cite{KITTI_RAW} using the split of Eigen et al. \cite{eigen2014depth}, with maximum depth set to 50m. Results concerned with state-of-the-art techniques for unsupervised monocular depth estimation leveraging video sequences (Temporal), binocular stereo pairs (Stereo) and both cues (Stereo+Temp.).}
		\label{table:eigen}
	\end{table*}
	
	\textbf{Smoothness Loss.} 
	This term favours the propagation of similar disparity values in low-textured regions, thus enforcing smoothness. It is obtained computing horizontal and vertical gradients on both disparity image and reference image, discouraging disparity smoothness in presence of strong image gradients.
	
	\begin{align}
	\begin{split}
	\mathcal{L}_{ds}(d,I) = \frac{1}{N} \sum_{ij} | \partial_x d_{ij}^l |e^{-|| \partial_x I_{ij}^l ||} + | \partial_y d_{ij}^l |e^{-|| \partial_y I_{ij}^l ||}
	\end{split}
	\label{eq:step2}
	\end{align}

	\textbf{Left-Right Disparity Consistency Loss}. 
	It enforces consistency between reference-to-target and target-to-reference disparity maps. It relies on the L1 distance between reference-to-target map and warped, according to the former, target-to-reference map.
	
	\begin{align}
	\begin{split}
	\mathcal{L}_{lr}(d^l,d^r) = \frac{1}{N} \sum_{ij} | d_{ij}^l - d_{ij + d_{ij}^l}^r|
	\end{split}
	\label{eq:step2}
	\end{align}

	\section{Depth computation and post-processing}
	\label{sec:pp}
	
	For the sake of clarity, we describe in detail how we combine $d^{cl}$ and $d^{cr}$ to obtain the final output map $d^c$. In \cite{godard2017unsupervised} 
	the authors obtained $d^l$ and $\hat{d^l}$ by processing, respectively, both $I$ and its horizontally flipped version $\hat{I}$. The two maps were combined as follows:
	
	\begin{equation}
	d_{pp} = \omega \cdot d^l + (1 - \omega) \cdot \hat{d^l}
	\end{equation}
	with $\omega$ obtained as:
	
	\begin{equation}
	\omega = \begin{cases} 0 &\textrm{if} j \leq 0.05 \\ 1 & \textrm{if} j > 0.95 \\ 0.5 & \textrm{otherwise}\\
	\end{cases}
	\label{eq:relu}
	\end{equation}
	
	being i, j normalized pixel coordinates.
	
	Following this principle, we combine our $d^{cl}$ and $d^{cr}$ maps as follows:
	
	\begin{equation}
	d^c = \omega \cdot d^{cr} + (1 - \omega) \cdot d^{cl} 
	\end{equation}
	
	Running two forwards, we can post-process both intermediate maps and 
	
	\begin{equation}
	d^c_{pp} = \omega \cdot d^{cr}_{pp} + (1 - \omega) \cdot d^{cl}_{pp} 
	\end{equation}
	
	being $d^{cr}_{pp}$ and $d^{cl}_{pp}$ obtained as:
	
	\begin{equation}
	d^{cr}_{pp} = \omega \cdot d^{cr} + (1 - \omega) \cdot \hat{d^{cr}} 
	\end{equation}
	
	\begin{equation}
	d^{cl}_{pp} = \omega \cdot \hat{d^{cl}} + (1 - \omega) \cdot d^{cl} 
	\end{equation}
	
	\section{Depth estimation: additional experiments with 50m cap}
	\label{sec:50m}
	
	We report additional experimental results on the Eigen split \cite{eigen2014depth}, evaluating depth maps up to a maximum distance of 50 meters as reported in some recent works \cite{godard2017unsupervised,zhan2018unsupervised,mahjourian2018unsupervised,zhou2017unsupervised}. Table \ref{table:eigen} contains a comparison 
	between all previous works reporting this experiment as well and our best model, i.e. 3Net ResNet50 + pp.
	This further evaluation confirms, once again, the superiority of our technique with respect to all competitors.
	
	\section{View synthesis and multi-baseline stereo}
	\label{sec:qualitative}
	
\begin{figure*}
		\centering
		
		\begin{overpic}[width=0.32\textwidth]{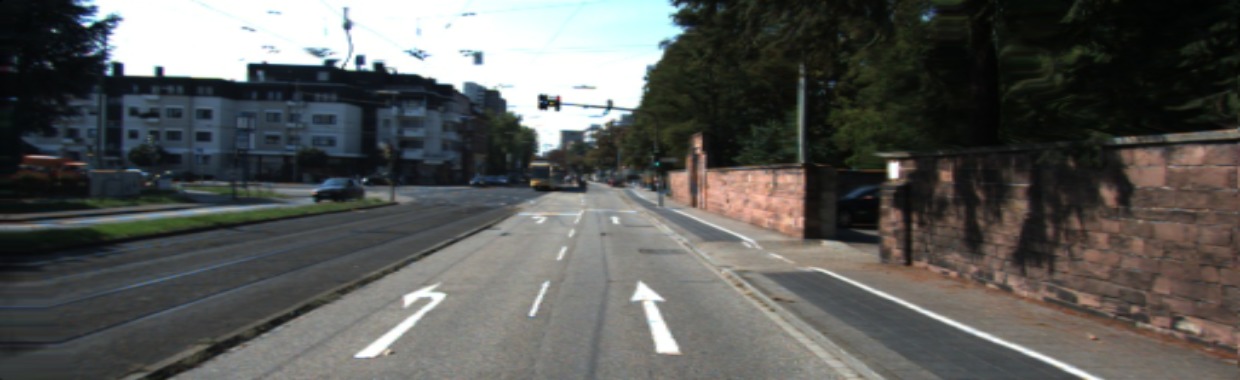}
			\put (2,4) {$\displaystyle\textcolor{white}{\textbf{(a)}}$}
		\end{overpic} 
		\begin{overpic}[width=0.32\textwidth]{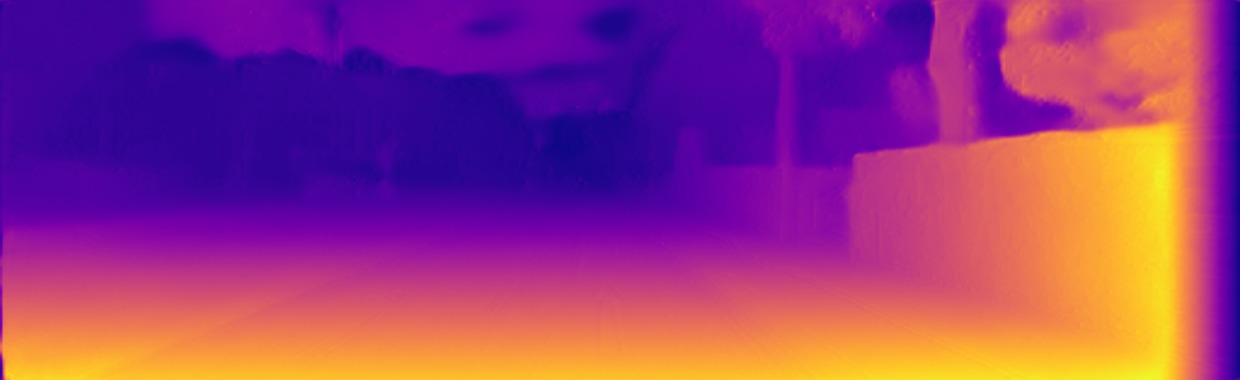}
			\put (2,4) {$\displaystyle\textcolor{white}{\textbf{(d)}}$}
		\end{overpic} 
		\begin{overpic}[width=0.32\textwidth]{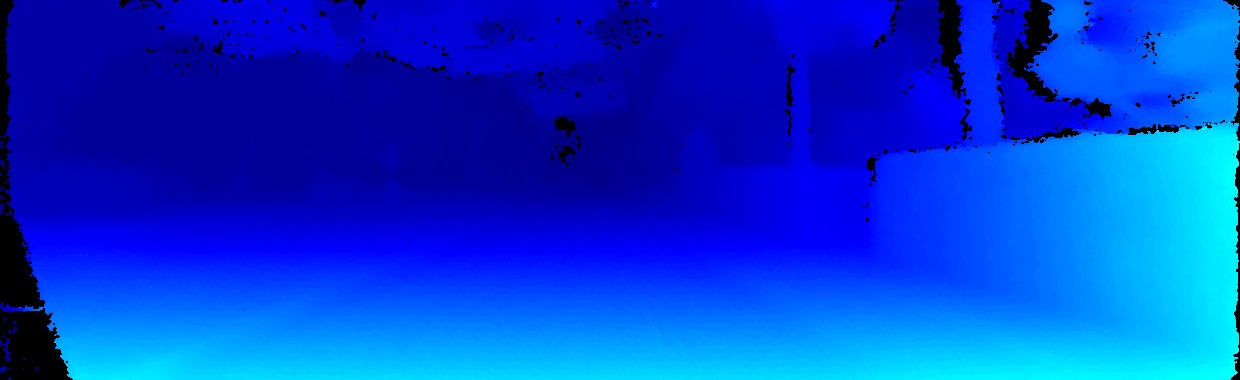}
			\put (2,4) {$\displaystyle\textcolor{white}{\textbf{(g)}}$}
		\end{overpic} 
		\\
		\begin{overpic}[width=0.32\textwidth]{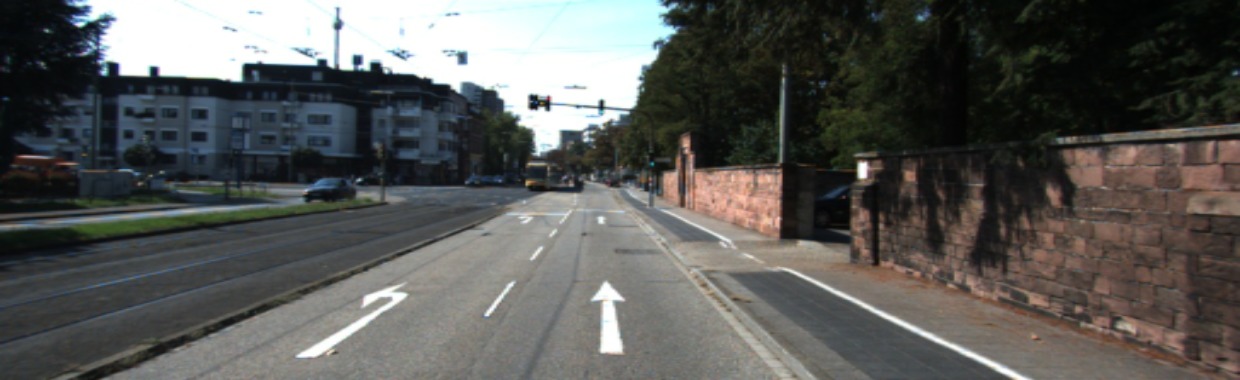}
			\put (2,4) {$\displaystyle\textcolor{white}{\textbf{(b)}}$}
		\end{overpic} 
		\begin{overpic}[width=0.32\textwidth]{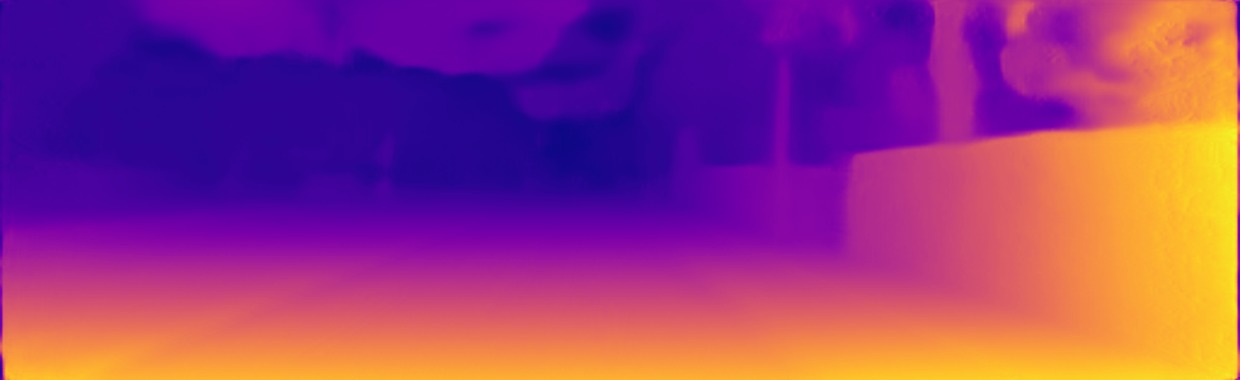}
			\put (2,4) {$\displaystyle\textcolor{white}{\textbf{(e)}}$}
		\end{overpic} 
		\begin{overpic}[width=0.32\textwidth]{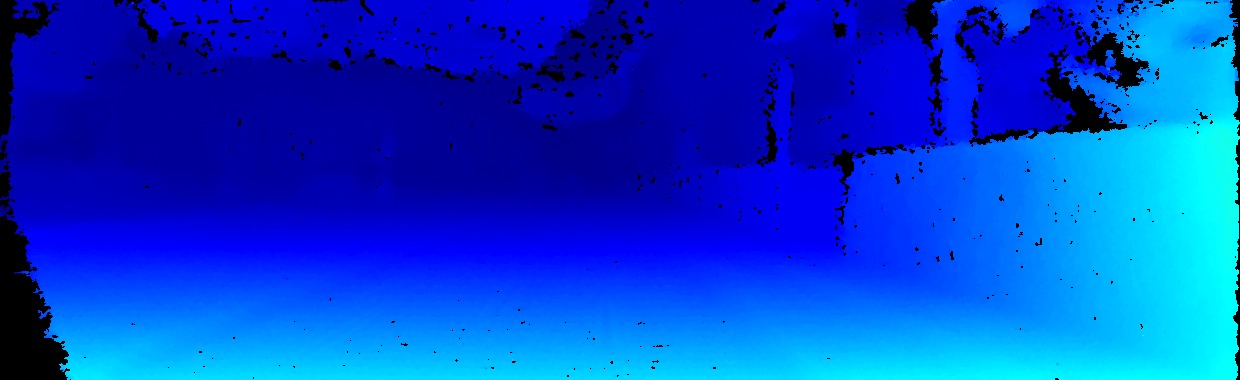}
			\put (2,4) {$\displaystyle\textcolor{white}{\textbf{(h)}}$}
		\end{overpic}
		\\
		\begin{overpic}[width=0.32\textwidth]{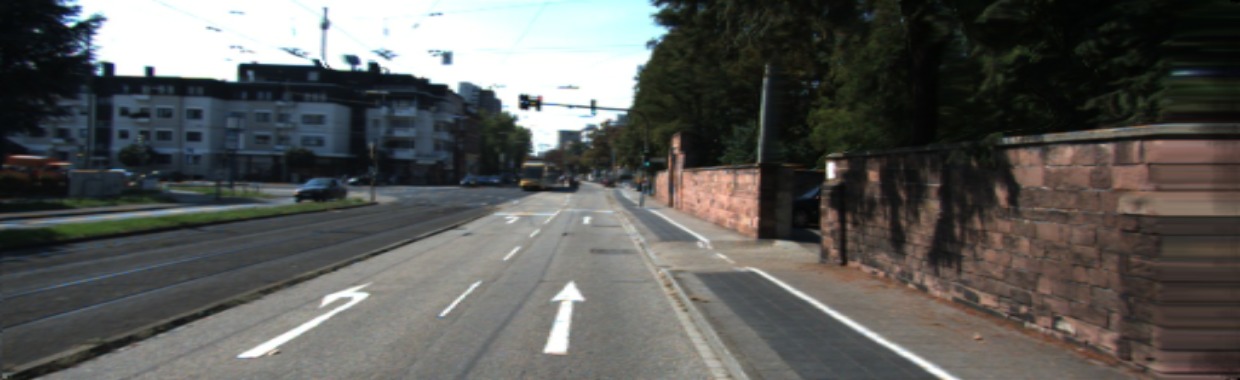}
			\put (2,4) {$\displaystyle\textcolor{white}{\textbf{(c)}}$}
		\end{overpic} 
		\begin{overpic}[width=0.32\textwidth]{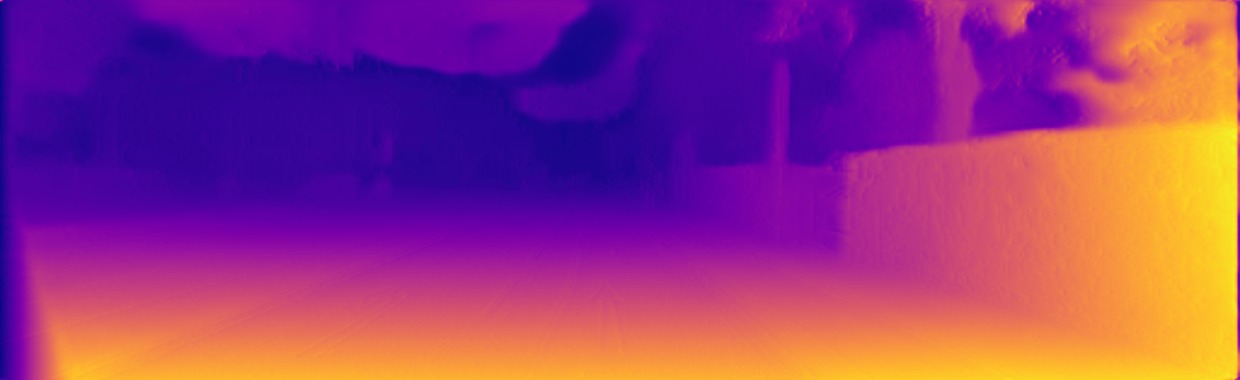}
			\put (2,4) {$\displaystyle\textcolor{white}{\textbf{(f)}}$}
		\end{overpic} 
		\begin{overpic}[width=0.32\textwidth]{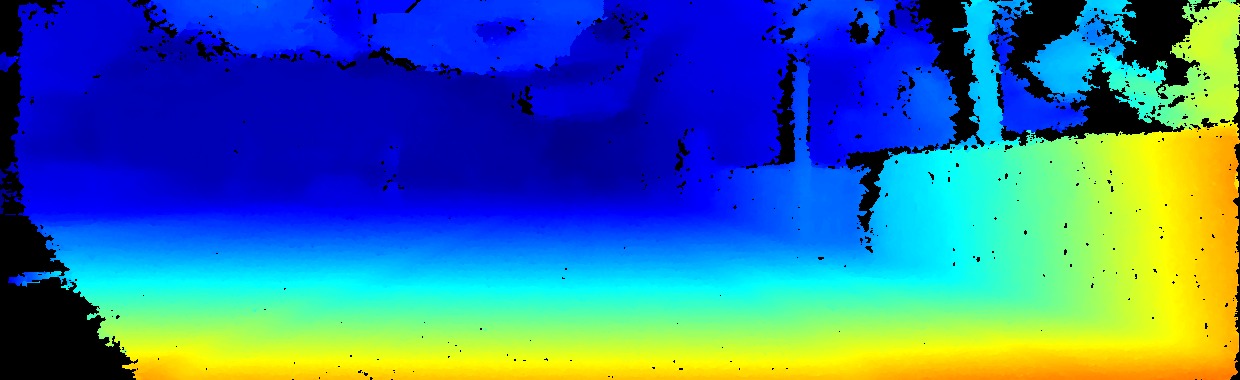}
			\put (2,4) {$\displaystyle\textcolor{white}{\textbf{(i)}}$}
		\end{overpic} 
		\\
		\hspace{1pt}
		\\
		\begin{overpic}[width=0.32\textwidth]{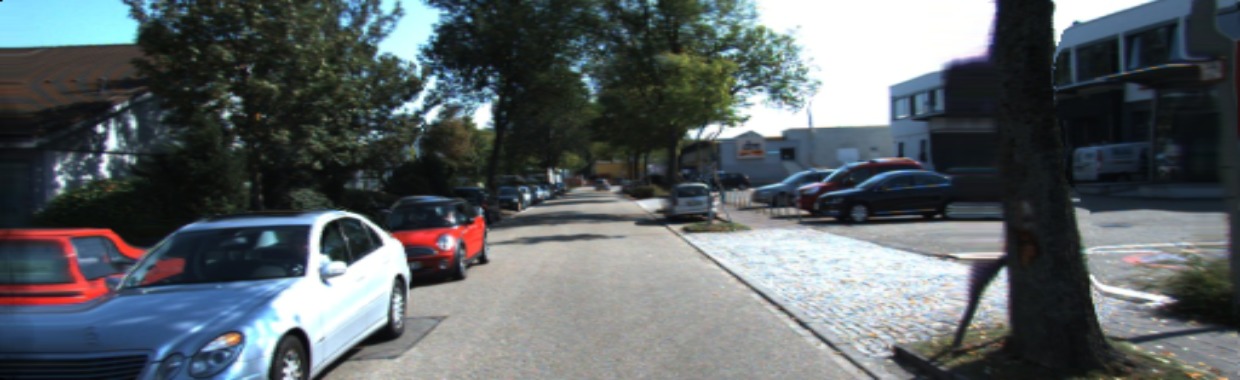}
			\put (2,4) {$\displaystyle\textcolor{white}{\textbf{(a)}}$}
		\end{overpic} 
		\begin{overpic}[width=0.32\textwidth]{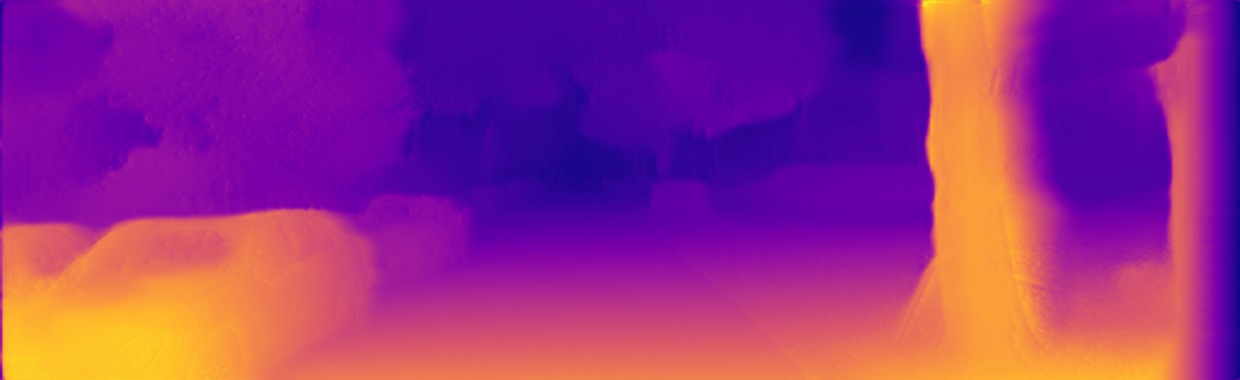}
			\put (2,4) {$\displaystyle\textcolor{white}{\textbf{(d)}}$}
		\end{overpic} 
		\begin{overpic}[width=0.32\textwidth]{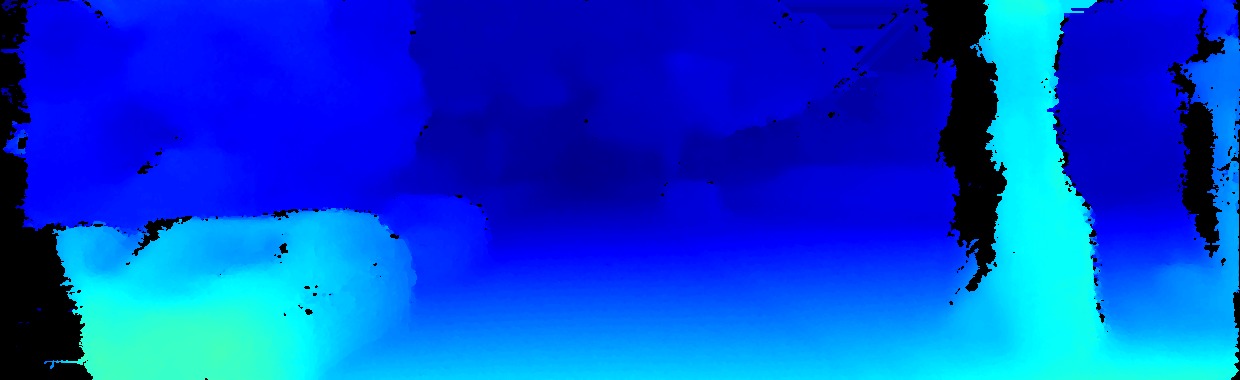}
			\put (2,4) {$\displaystyle\textcolor{white}{\textbf{(g)}}$}
		\end{overpic} 
		\\
		\begin{overpic}[width=0.32\textwidth]{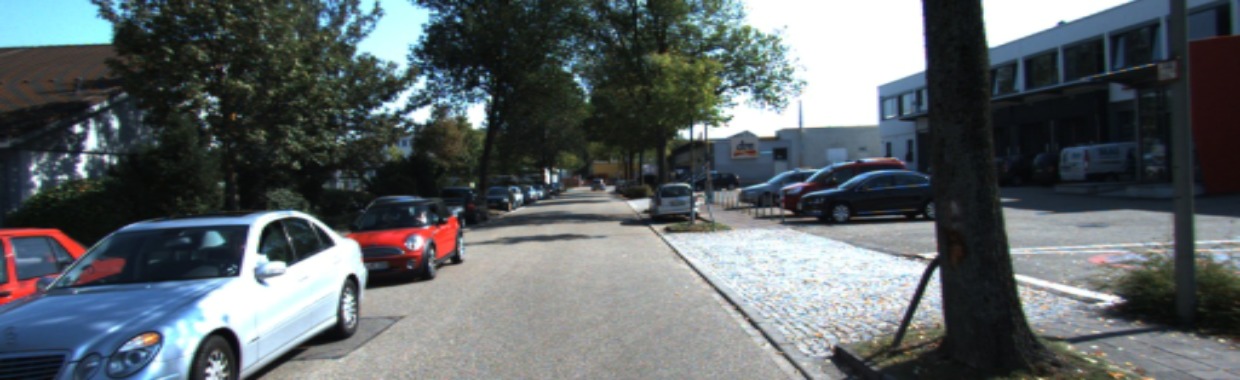}
			\put (2,4) {$\displaystyle\textcolor{white}{\textbf{(b)}}$}
		\end{overpic} 
		\begin{overpic}[width=0.32\textwidth]{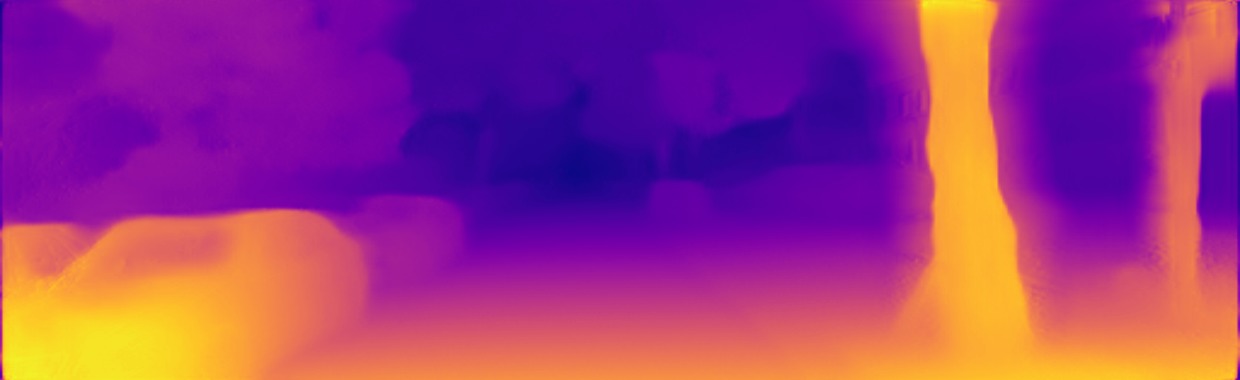}
			\put (2,4) {$\displaystyle\textcolor{white}{\textbf{(e)}}$}
		\end{overpic} 
		\begin{overpic}[width=0.32\textwidth]{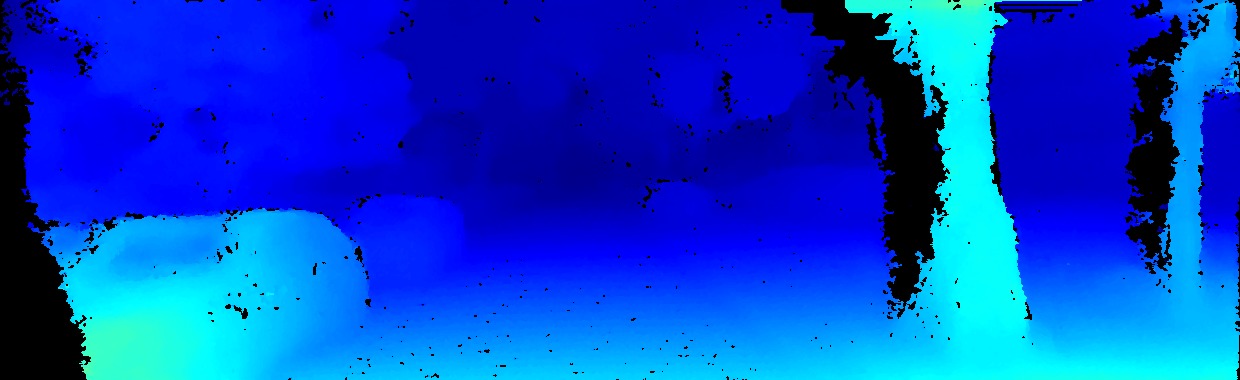}
			\put (2,4) {$\displaystyle\textcolor{white}{\textbf{(h)}}$}
		\end{overpic}
		\\
		\begin{overpic}[width=0.32\textwidth]{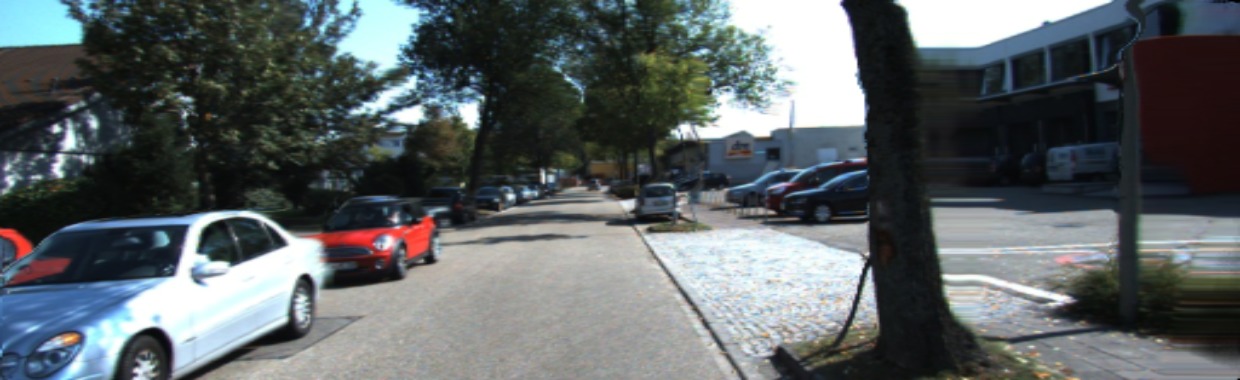}
			\put (2,4) {$\displaystyle\textcolor{white}{\textbf{(c)}}$}
		\end{overpic} 
		\begin{overpic}[width=0.32\textwidth]{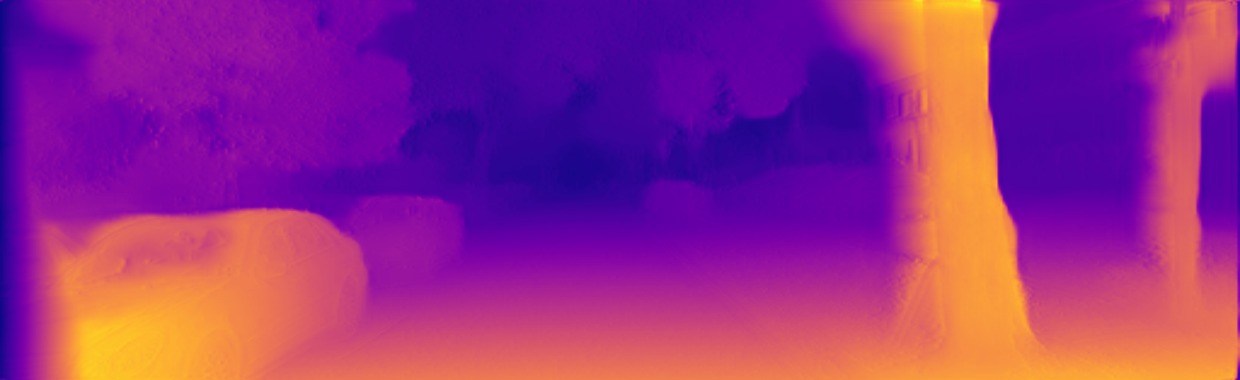}
			\put (2,4) {$\displaystyle\textcolor{white}{\textbf{(f)}}$}
		\end{overpic} 
		\begin{overpic}[width=0.32\textwidth]{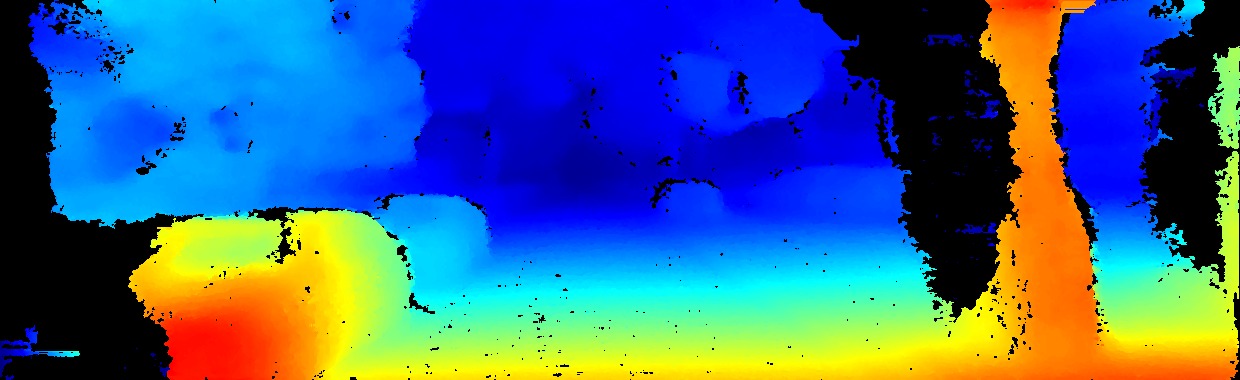}
			\put (2,4) {$\displaystyle\textcolor{white}{\textbf{(i)}}$}
		\end{overpic} 
		\\
		\hspace{1pt}
		\\
		\begin{overpic}[width=0.32\textwidth]{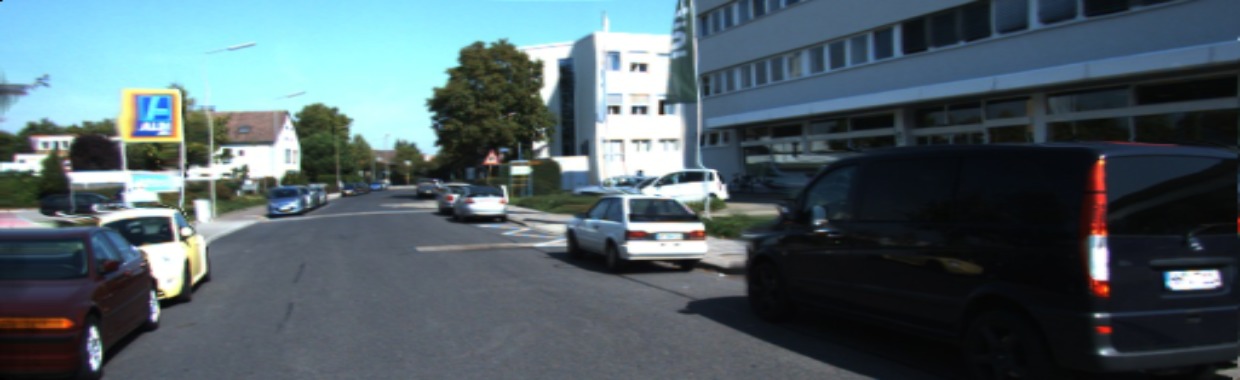}
			\put (2,4) {$\displaystyle\textcolor{white}{\textbf{(a)}}$}
		\end{overpic} 
		\begin{overpic}[width=0.32\textwidth]{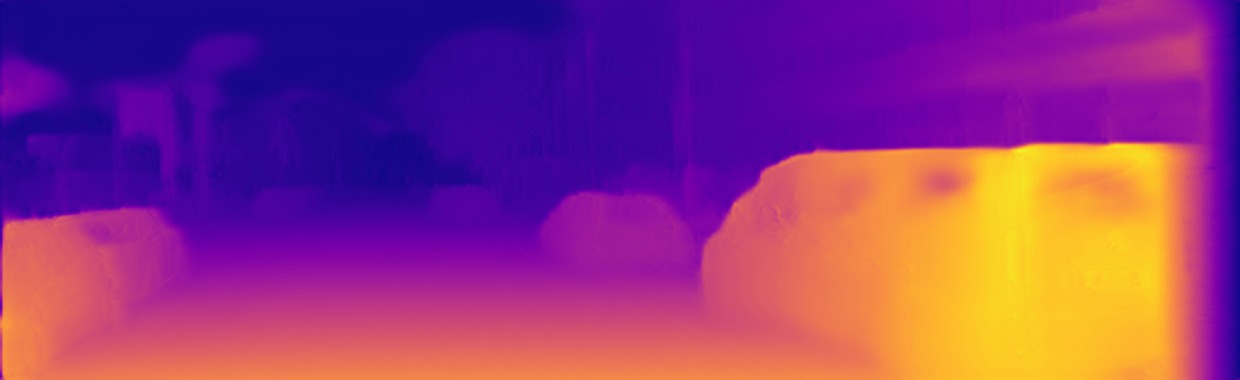}
			\put (2,4) {$\displaystyle\textcolor{white}{\textbf{(d)}}$}
		\end{overpic} 
		\begin{overpic}[width=0.32\textwidth]{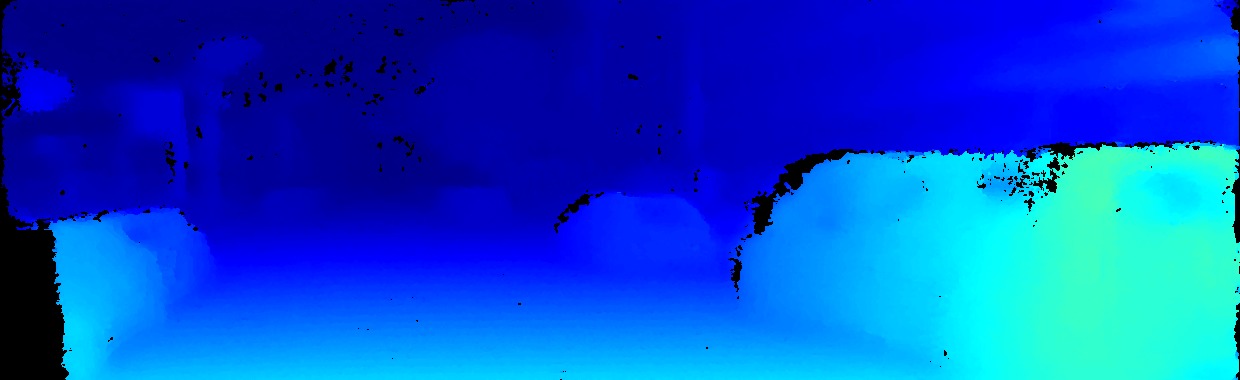}
			\put (2,4) {$\displaystyle\textcolor{white}{\textbf{(g)}}$}
		\end{overpic} 
		\\
		\begin{overpic}[width=0.32\textwidth]{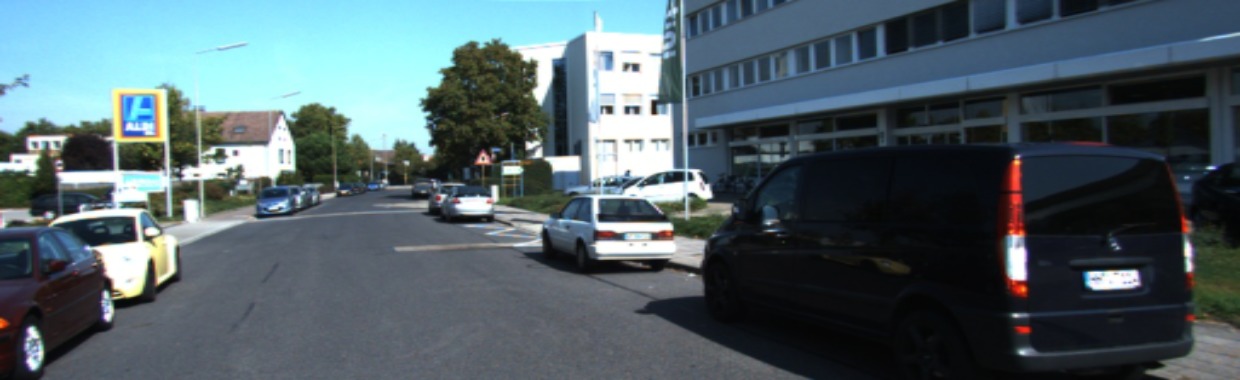}
			\put (2,4) {$\displaystyle\textcolor{white}{\textbf{(b)}}$}
		\end{overpic} 
		\begin{overpic}[width=0.32\textwidth]{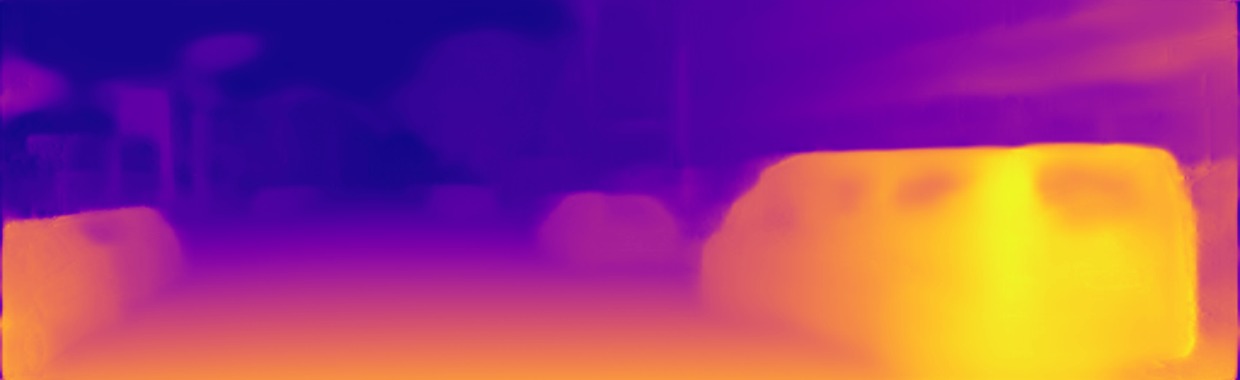}
			\put (2,4) {$\displaystyle\textcolor{white}{\textbf{(e)}}$}
		\end{overpic} 
		\begin{overpic}[width=0.32\textwidth]{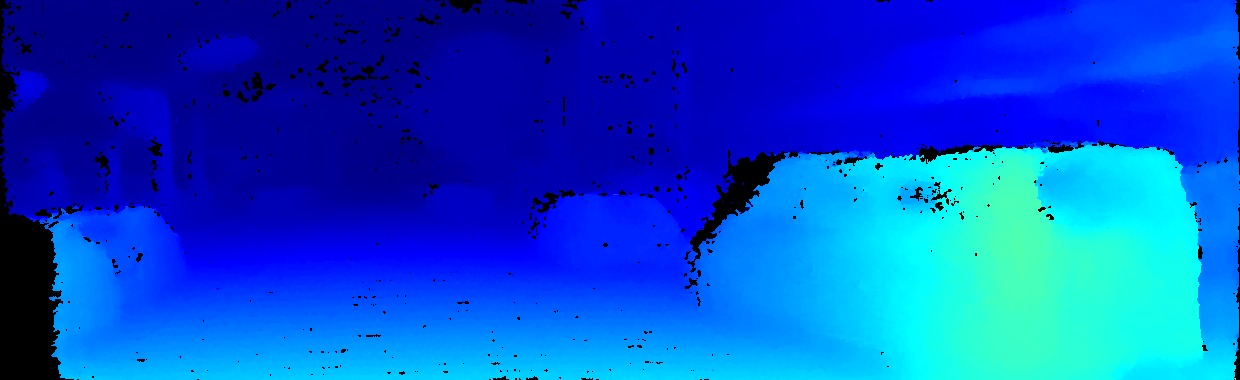}
			\put (2,4) {$\displaystyle\textcolor{white}{\textbf{(h)}}$}
		\end{overpic}
		\\
		\begin{overpic}[width=0.32\textwidth]{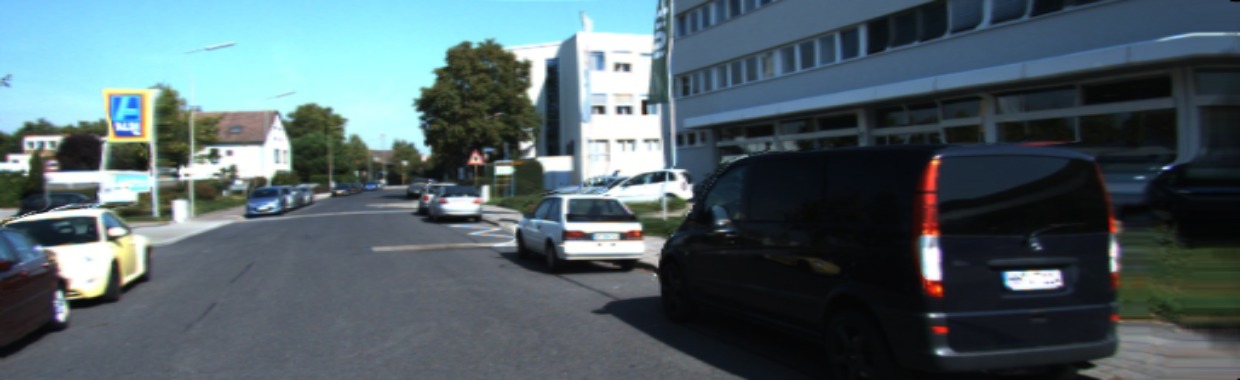}
			\put (2,4) {$\displaystyle\textcolor{white}{\textbf{(c)}}$}
		\end{overpic} 
		\begin{overpic}[width=0.32\textwidth]{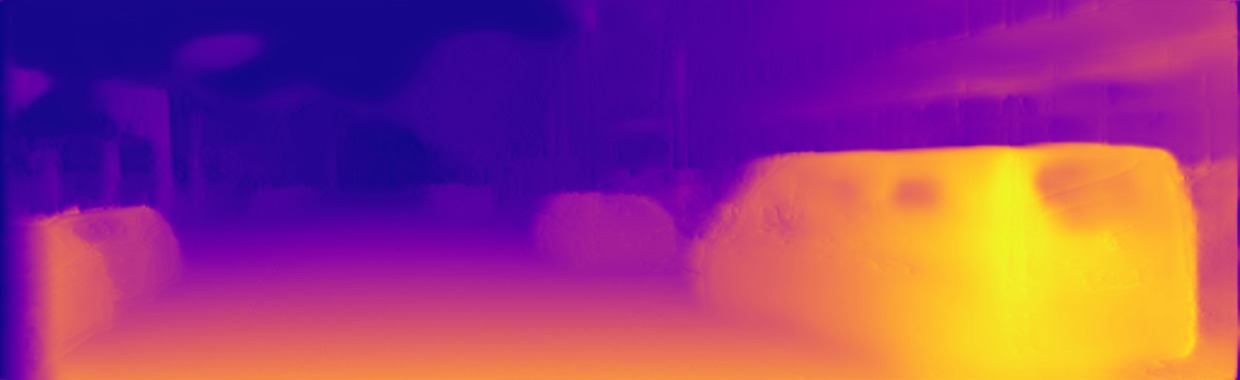}
			\put (2,4) {$\displaystyle\textcolor{white}{\textbf{(f)}}$}
		\end{overpic} 
		\begin{overpic}[width=0.32\textwidth]{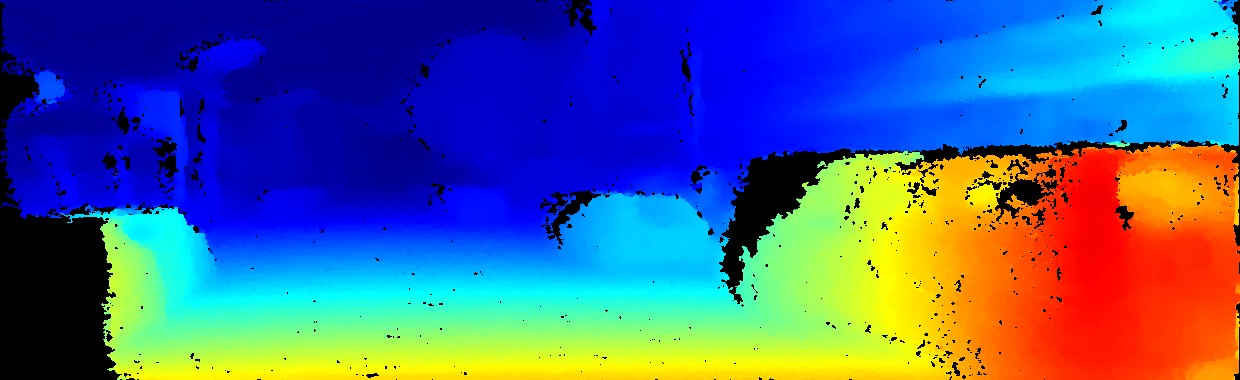}
			\put (2,4) {$\displaystyle\textcolor{white}{\textbf{(i)}}$}
		\end{overpic} 
		\\    
		\caption{Qualitative evaluation of 3Net. In the leftmost column, we show (always from top to bottom) \emph{synthetic} left  (a), \emph{real} central (b) and \emph{synthetic} right (c) view. In the middle column, $d^{cl}$ (d), $d^c$ (e) and $d^{cr}$ (f) depth maps computed by our network processing the input image. In the rightmost column, disparity maps obtained by the SGM algorithm \cite{hirschmuller2005accurate} processing respectively, left-center (g), center-right (h) and left-right (i) stereo pair.}
		\label{fig:q1}
	\end{figure*}

\begin{figure*}
		\centering
		
		\begin{overpic}[width=0.32\textwidth]{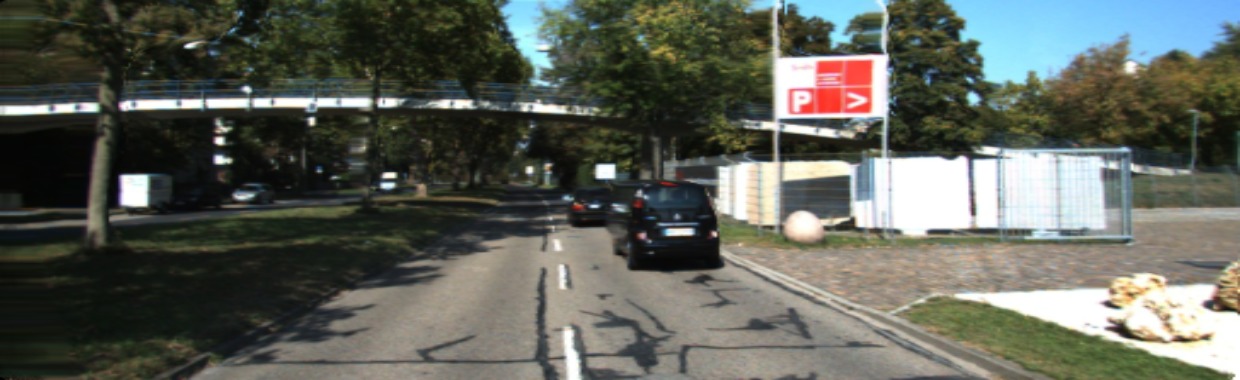}
			\put (2,4) {$\displaystyle\textcolor{white}{\textbf{(a)}}$}
		\end{overpic} 
		\begin{overpic}[width=0.32\textwidth]{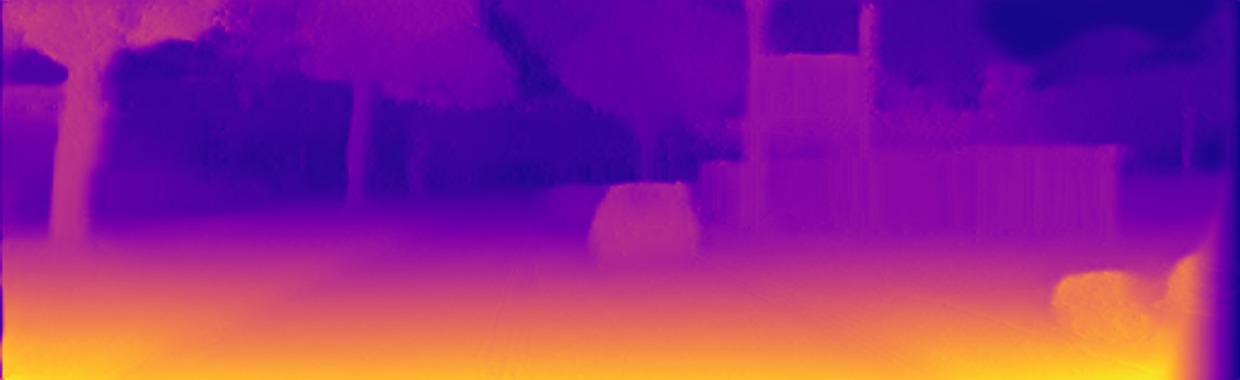}
			\put (2,4) {$\displaystyle\textcolor{white}{\textbf{(d)}}$}
		\end{overpic} 
		\begin{overpic}[width=0.32\textwidth]{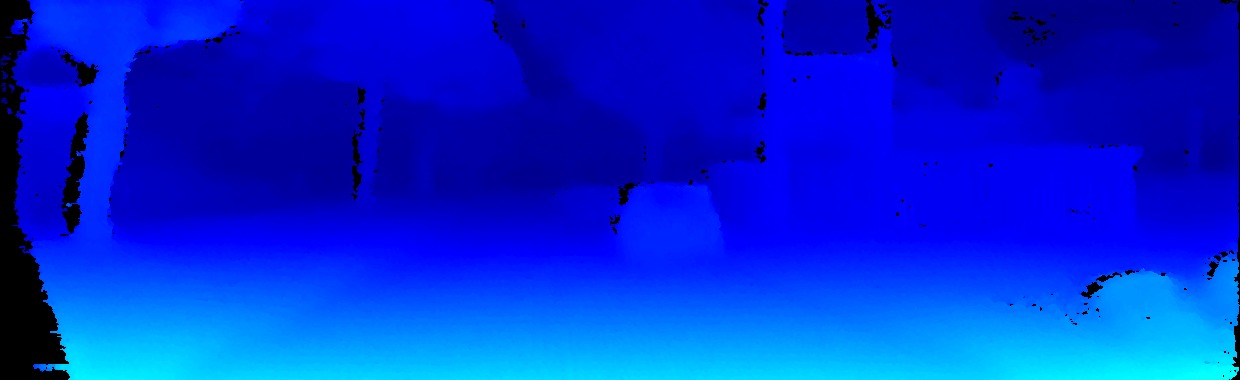}
			\put (2,4) {$\displaystyle\textcolor{white}{\textbf{(g)}}$}
		\end{overpic} 
		\\
		\begin{overpic}[width=0.32\textwidth]{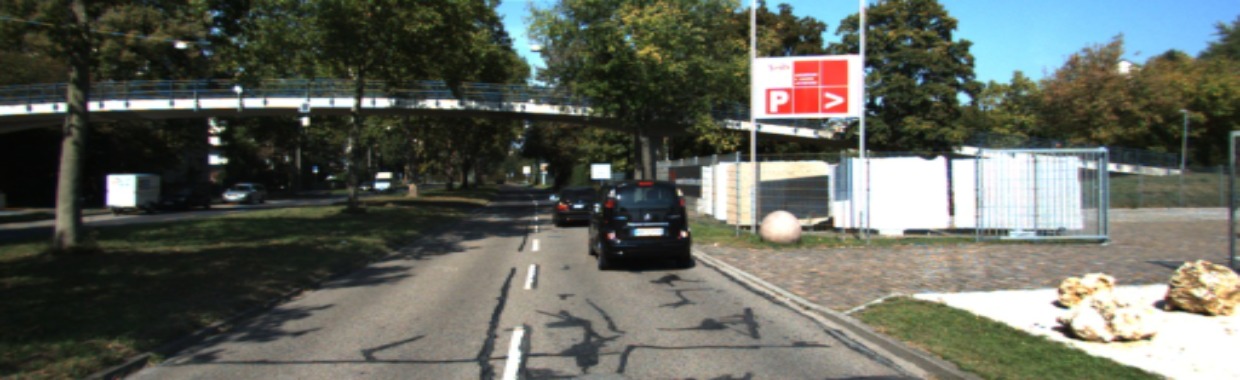}
			\put (2,4) {$\displaystyle\textcolor{white}{\textbf{(b)}}$}
		\end{overpic} 
		\begin{overpic}[width=0.32\textwidth]{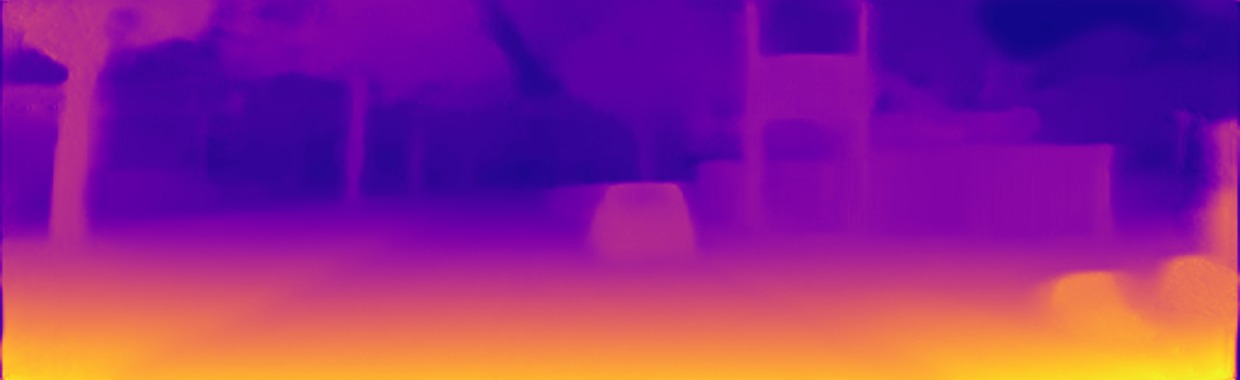}
			\put (2,4) {$\displaystyle\textcolor{white}{\textbf{(e)}}$}
		\end{overpic} 
		\begin{overpic}[width=0.32\textwidth]{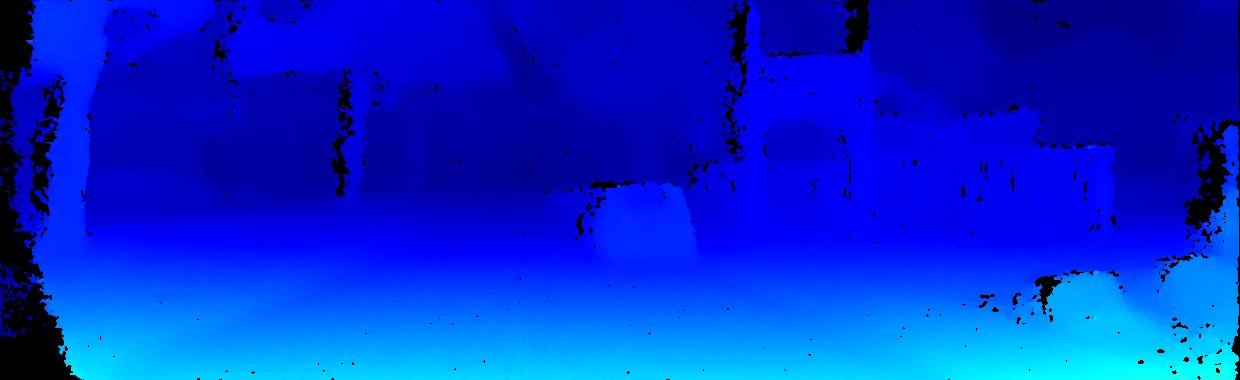}
			\put (2,4) {$\displaystyle\textcolor{white}{\textbf{(h)}}$}
		\end{overpic}
		\\
		\begin{overpic}[width=0.32\textwidth]{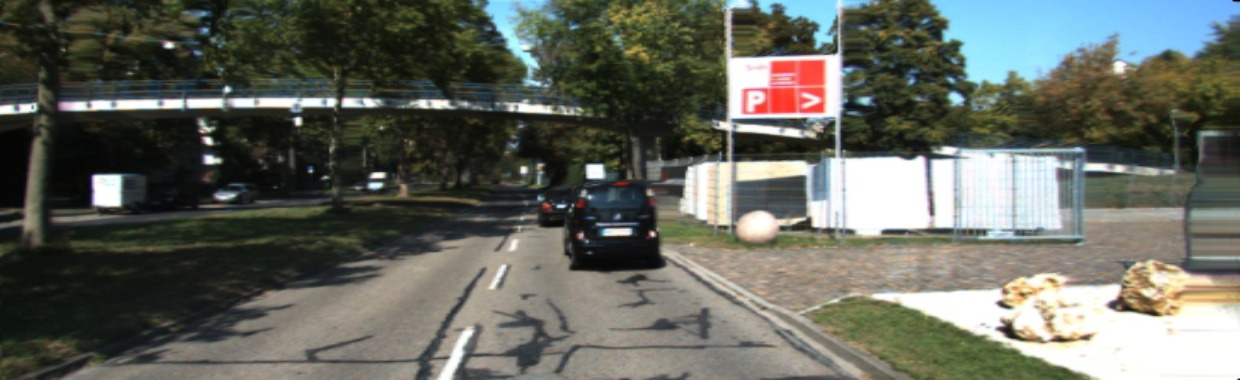}
			\put (2,4) {$\displaystyle\textcolor{white}{\textbf{(c)}}$}
		\end{overpic} 
		\begin{overpic}[width=0.32\textwidth]{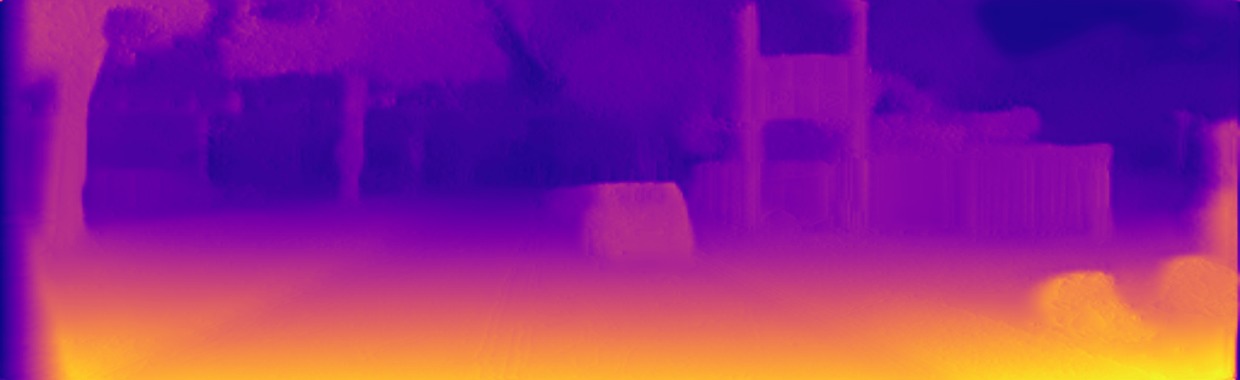}
			\put (2,4) {$\displaystyle\textcolor{white}{\textbf{(f)}}$}
		\end{overpic} 
		\begin{overpic}[width=0.32\textwidth]{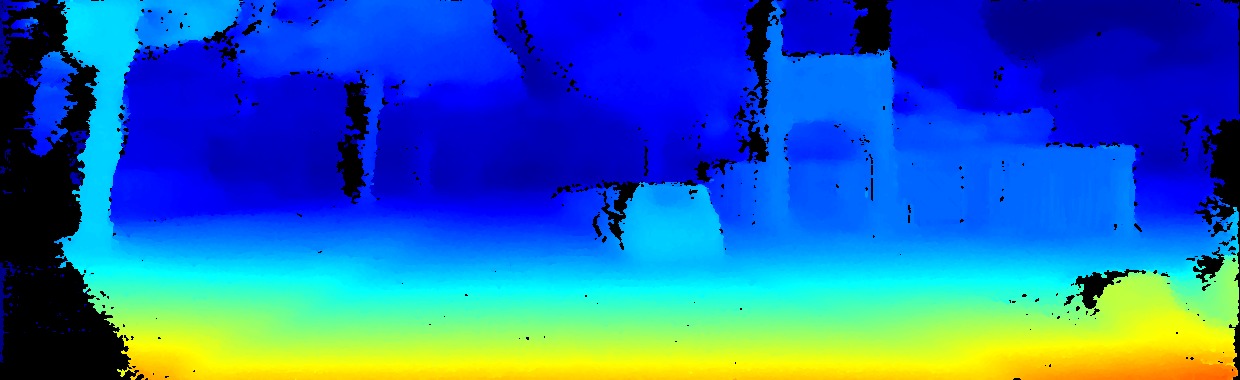}
			\put (2,4) {$\displaystyle\textcolor{white}{\textbf{(i)}}$}
		\end{overpic} 
		\\
		\hspace{1pt}
		\\
		\begin{overpic}[width=0.32\textwidth]{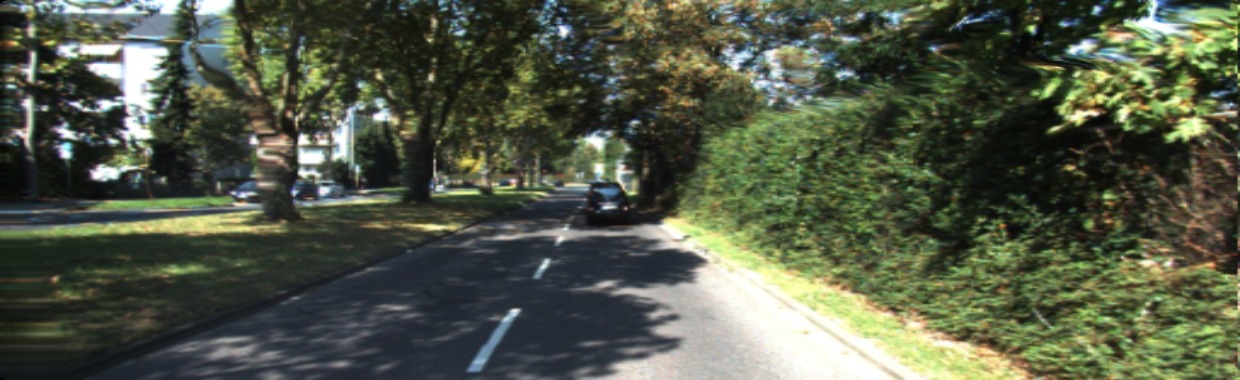}
			\put (2,4) {$\displaystyle\textcolor{white}{\textbf{(a)}}$}
		\end{overpic} 
		\begin{overpic}[width=0.32\textwidth]{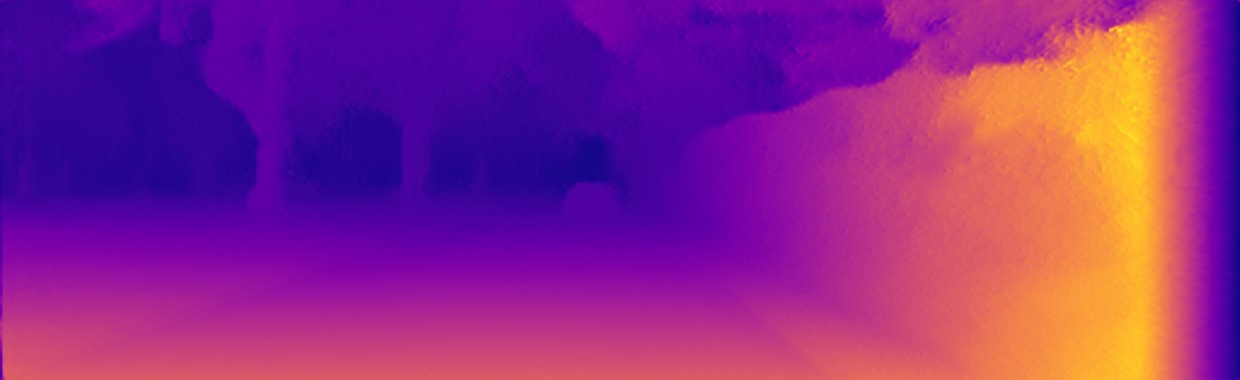}
			\put (2,4) {$\displaystyle\textcolor{white}{\textbf{(d)}}$}
		\end{overpic} 
		\begin{overpic}[width=0.32\textwidth]{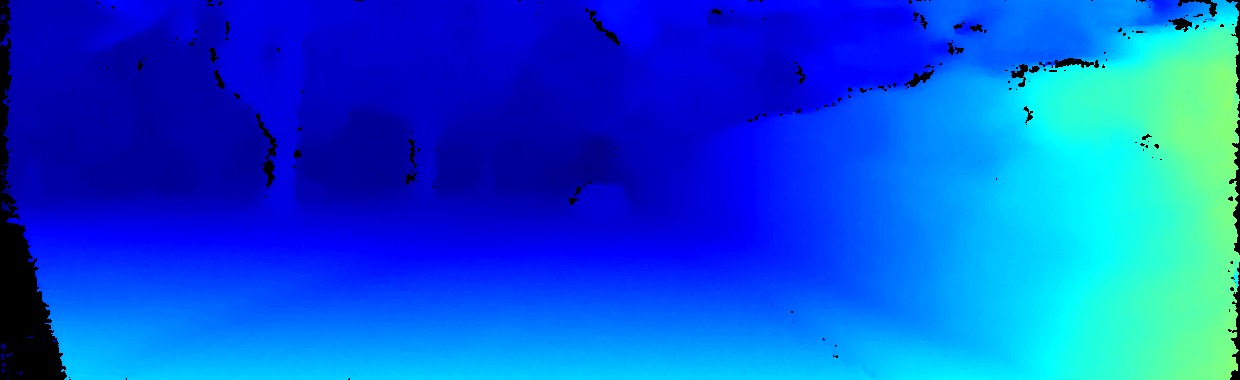}
			\put (2,4) {$\displaystyle\textcolor{white}{\textbf{(g)}}$}
		\end{overpic} 
		\\
		\begin{overpic}[width=0.32\textwidth]{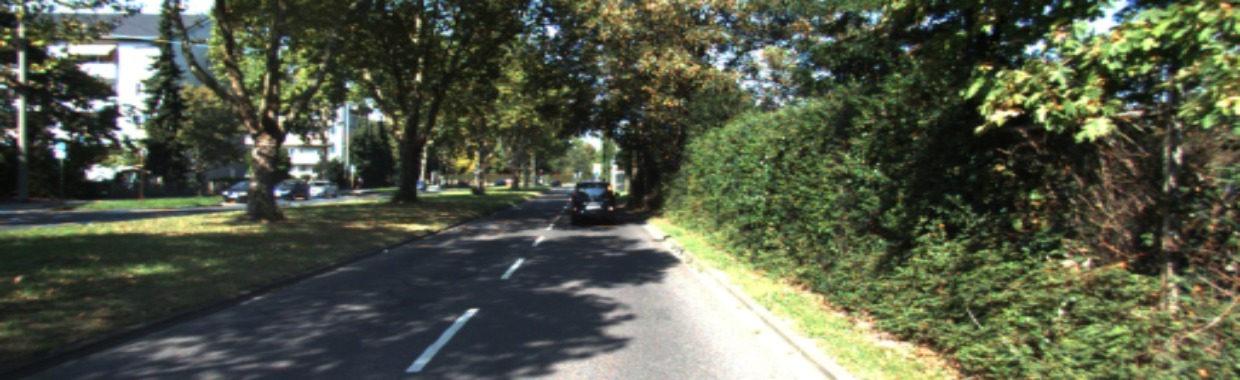}
			\put (2,4) {$\displaystyle\textcolor{white}{\textbf{(b)}}$}
		\end{overpic} 
		\begin{overpic}[width=0.32\textwidth]{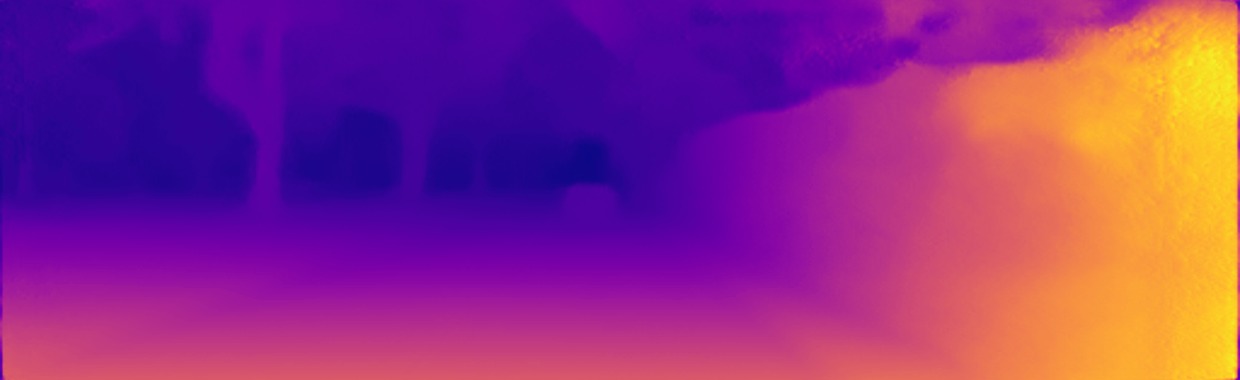}
			\put (2,4) {$\displaystyle\textcolor{white}{\textbf{(e)}}$}
		\end{overpic} 
		\begin{overpic}[width=0.32\textwidth]{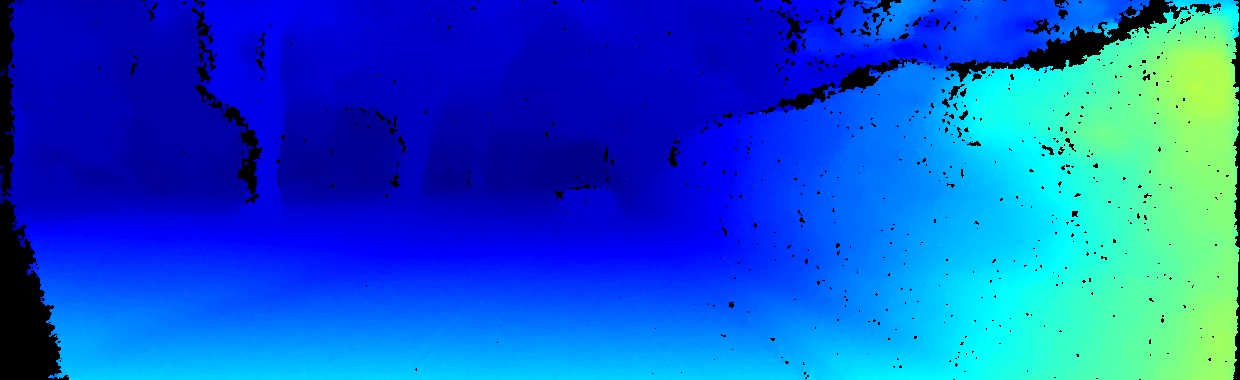}
			\put (2,4) {$\displaystyle\textcolor{white}{\textbf{(h)}}$}
		\end{overpic}
		\\
		\begin{overpic}[width=0.32\textwidth]{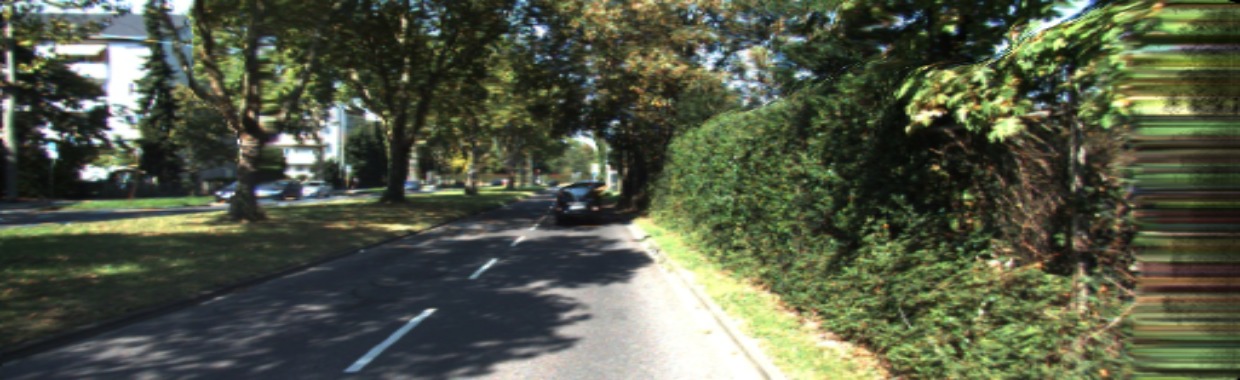}
			\put (2,4) {$\displaystyle\textcolor{white}{\textbf{(c)}}$}
		\end{overpic} 
		\begin{overpic}[width=0.32\textwidth]{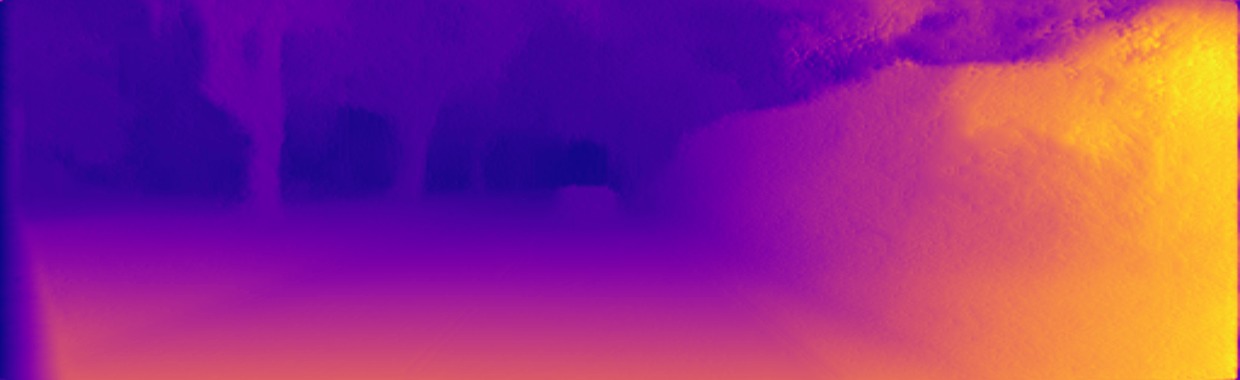}
			\put (2,4) {$\displaystyle\textcolor{white}{\textbf{(f)}}$}
		\end{overpic} 
		\begin{overpic}[width=0.32\textwidth]{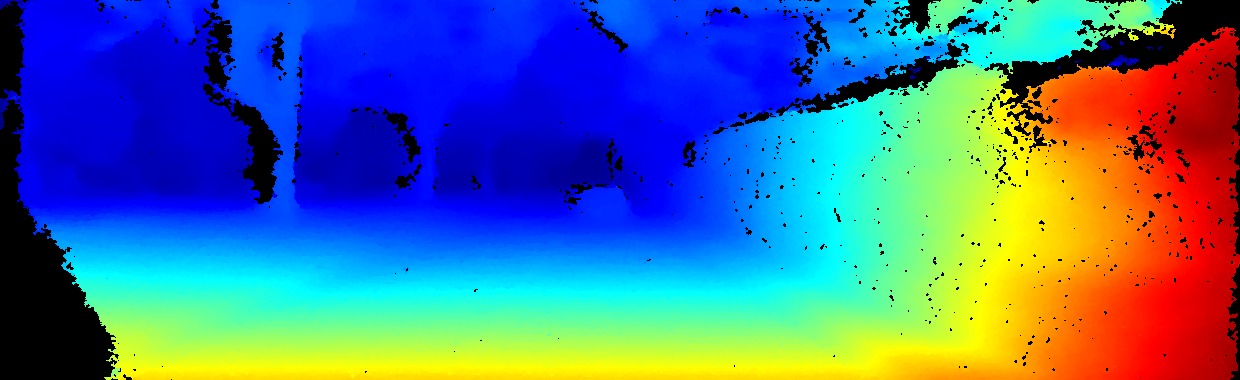}
			\put (2,4) {$\displaystyle\textcolor{white}{\textbf{(i)}}$}
		\end{overpic} 
		\\
		\hspace{1pt}
		\\
		\begin{overpic}[width=0.32\textwidth]{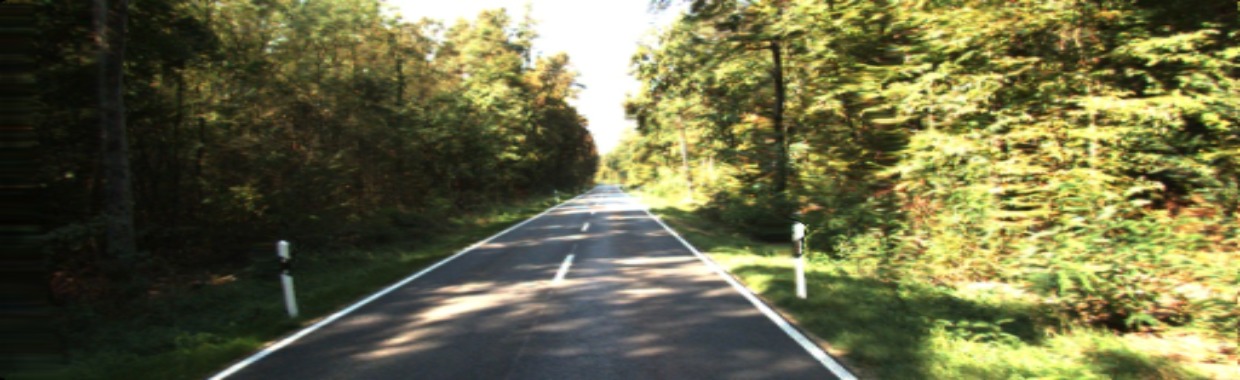}
			\put (2,4) {$\displaystyle\textcolor{white}{\textbf{(a)}}$}
		\end{overpic} 
		\begin{overpic}[width=0.32\textwidth]{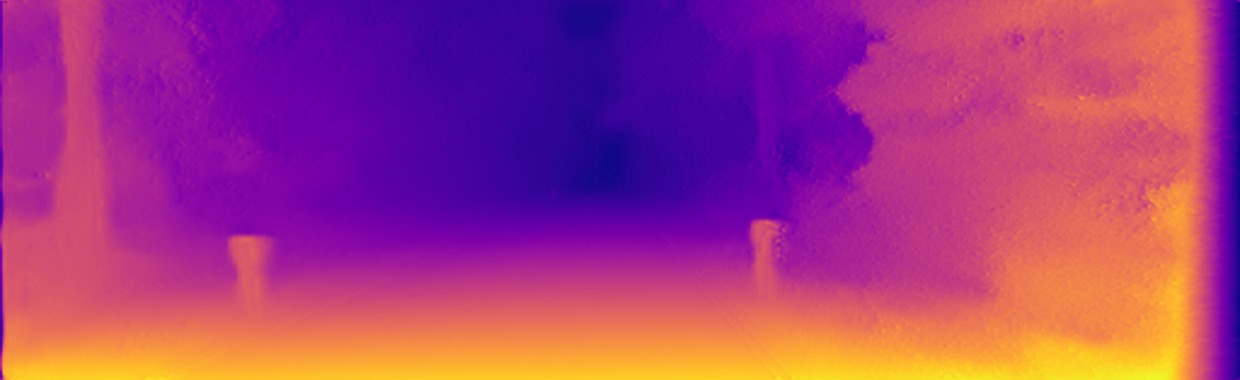}
			\put (2,4) {$\displaystyle\textcolor{white}{\textbf{(d)}}$}
		\end{overpic} 
		\begin{overpic}[width=0.32\textwidth]{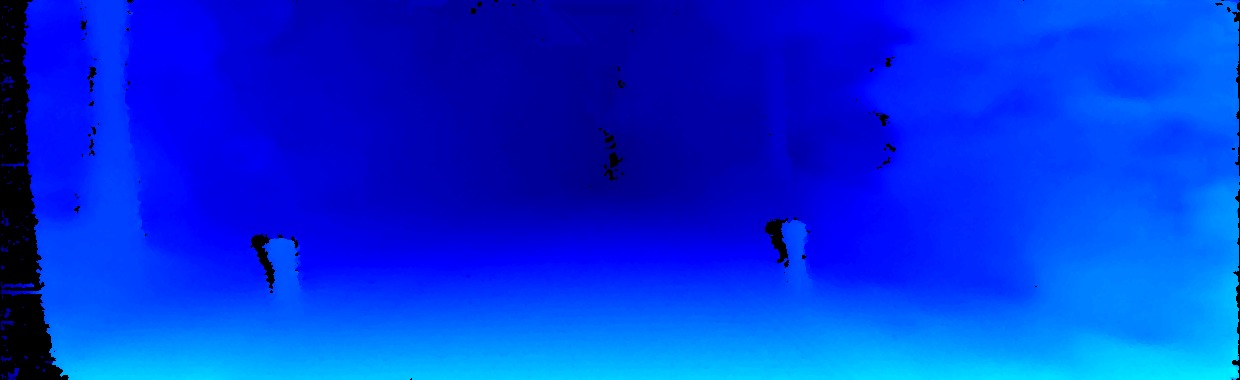}
			\put (2,4) {$\displaystyle\textcolor{white}{\textbf{(g)}}$}
		\end{overpic} 
		\\
		\begin{overpic}[width=0.32\textwidth]{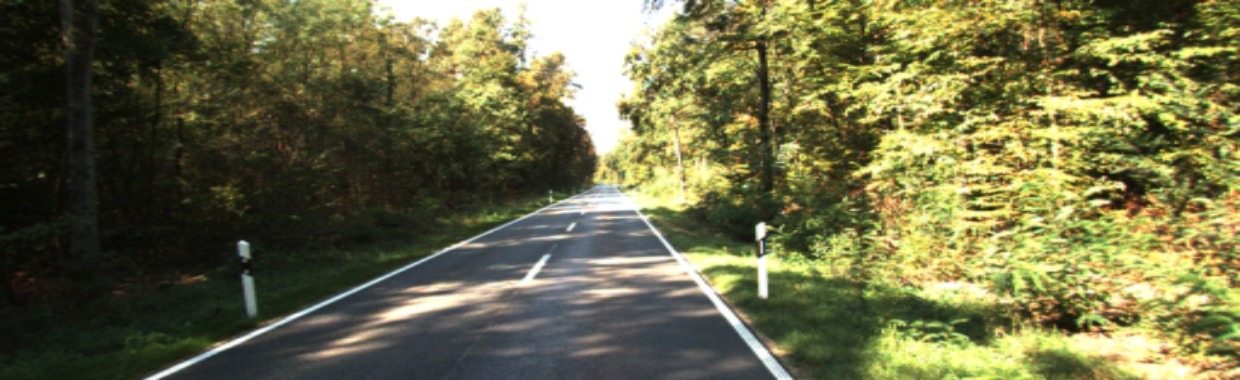}
			\put (2,4) {$\displaystyle\textcolor{white}{\textbf{(b)}}$}
		\end{overpic} 
		\begin{overpic}[width=0.32\textwidth]{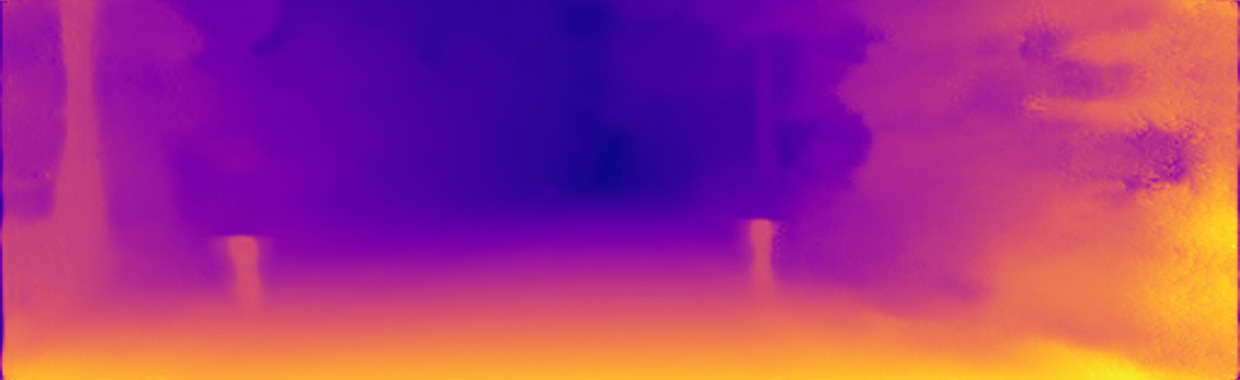}
			\put (2,4) {$\displaystyle\textcolor{white}{\textbf{(e)}}$}
		\end{overpic} 
		\begin{overpic}[width=0.32\textwidth]{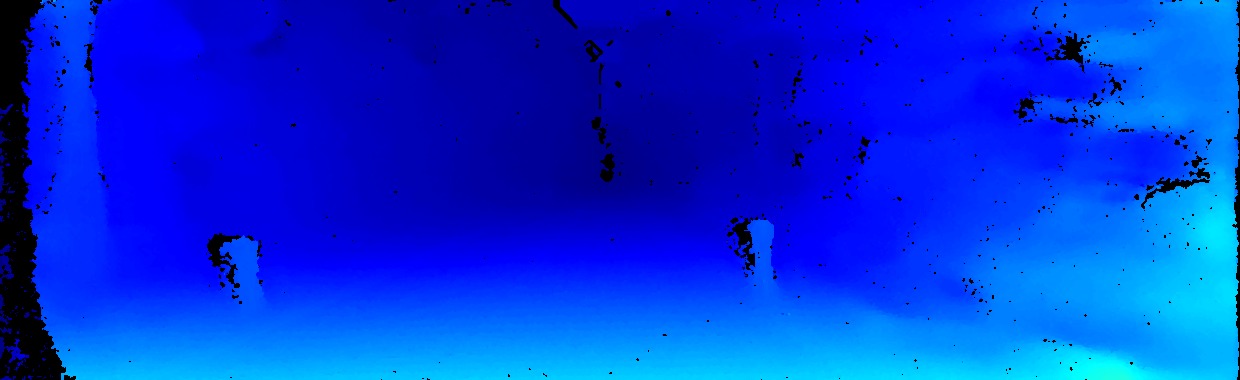}
			\put (2,4) {$\displaystyle\textcolor{white}{\textbf{(h)}}$}
		\end{overpic}
		\\
		\begin{overpic}[width=0.32\textwidth]{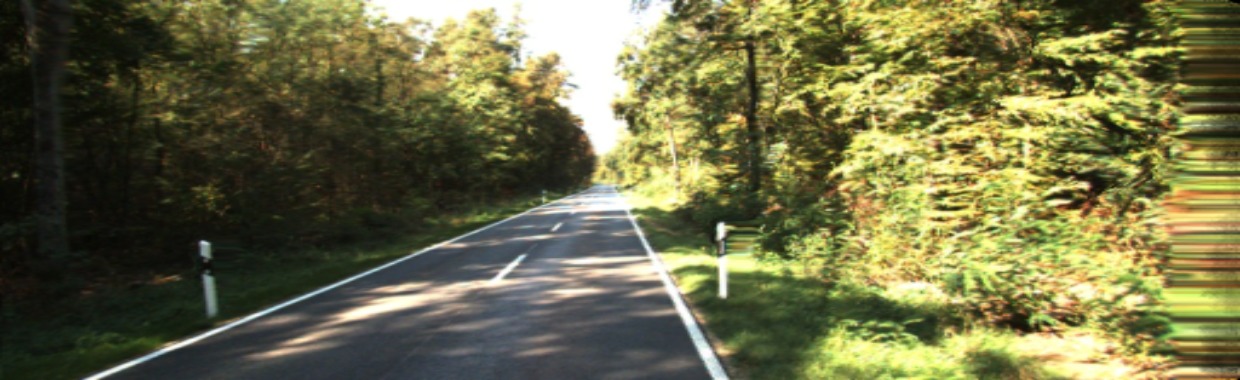}
			\put (2,4) {$\displaystyle\textcolor{white}{\textbf{(c)}}$}
		\end{overpic} 
		\begin{overpic}[width=0.32\textwidth]{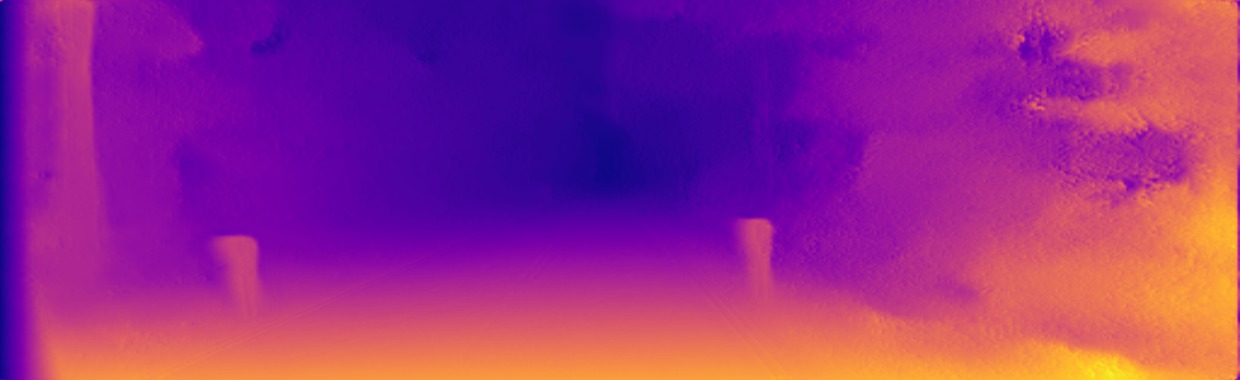}
			\put (2,4) {$\displaystyle\textcolor{white}{\textbf{(f)}}$}
		\end{overpic} 
		\begin{overpic}[width=0.32\textwidth]{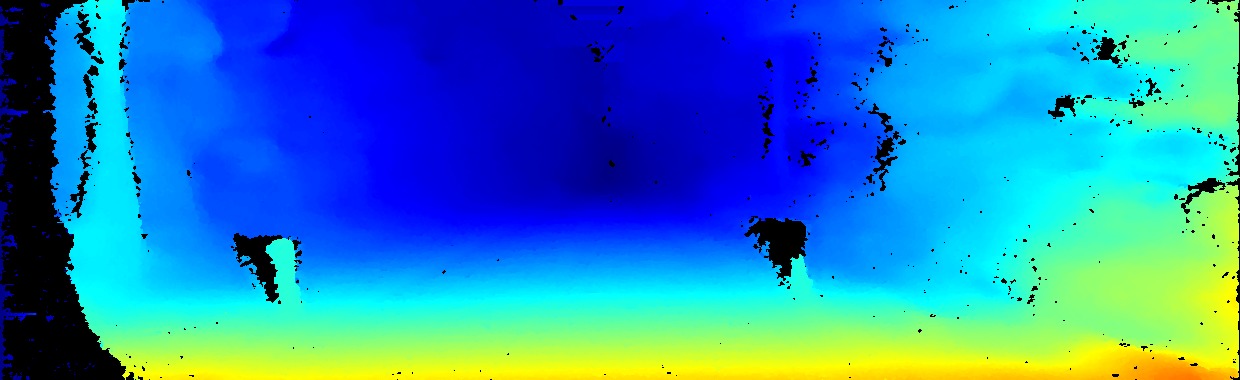}
			\put (2,4) {$\displaystyle\textcolor{white}{\textbf{(i)}}$}
		\end{overpic} 
		\\    
		\caption{Qualitative evaluation of 3Net. In the leftmost column, we show (always from top to bottom) \emph{synthetic} left  (a), \emph{real} central (b) and \emph{synthetic} right (c) view. In the middle column, $d^{cl}$ (d), $d^c$ (e) and $d^{cr}$ (f) depth maps computed by our network processing the input image. In the rightmost column, disparity maps obtained with SGM algorithm \cite{hirschmuller2005accurate} processing respectively, left-center (g), center-right (h) and left-right (i) stereo pair.}
		\label{fig:q2}
	\end{figure*}

	Finally, deploying 3Net ResNet50 + pp trained on CS+E, we provide additional qualitative results for depth-from-mono estimation and view synthesis. Figure \ref{fig:q1} and \ref{fig:q2} reports six examples taken from the evaluation set of the Eigen split \cite{eigen2014depth}. In particular, we show in the leftmost column the generated left view (a), the single input image fed to our network (b) and the generated right view (c). In the mid column, the three output maps of 3Net, respectively, $d^{cl}$ (d), $d^{c}$ (e) and $d^{cr}$ (f). Finally, in the rightmost column, we report disparity maps obtained processing with SGM \cite{hirschmuller2005accurate} the three stereo pairs obtainable with 3Net from the three views (one real, two synthetic) depicted in the leftmost column. In particular, the disparity maps computed by SGM are concerned with three stereo pairs: left-to-center (g), center-to-right (h) and left-to-right (i). It is worth to note that the left-to-right stereo pair (i) is made of two completely novel views synthesized by our network. The other two stereo pairs contain the input image and a a novel image synthesized by 3Net. 
	
	Observing (a), (b) and (c) we can easily notice three different view points: the two \emph{virtual} cameras are located at the left and right side of the \emph{real} camera (i.e., the central one). The three maps in the middle column clearly show artifacts occurring near depth discontinuities and occlusions in (d) and (f) and how they are greatly dampen in the final output of our network (e).
	Finally, we can perceive how (g) and (i) share the same reference image (synthetic left) and how they compute different disparity values according to different baselines, \emph{narrow} and \emph{wide}, made available by the three-view \emph{virtual} rig enabled by 3Net.

	A video showing the performance of 3Net on the KITTI sequence \emph{2011\_10\_03\_drive\_0047\_sync} \cite{KITTI_RAW} not part of the Eigen split imagery used for training is available at this link: \url{https://www.youtube.com/watch?v=uMA5YWJME4M}.\\
	Finally, the source code is available at this link: \url{https://github.com/mattpoggi/3net}

	\begin{table}
		\centering
		\begin{tabular}{|c|cc|cc|}
			\hline
			& \multicolumn{2}{c|}{256$\times$512} & \multicolumn{2}{c|}{384$\times$1280} \\
			& 1$\times$ & 2$\times$ & 1$\times$ & 2$\times$ \\
			\hline
			\cite{godard2017unsupervised} ResNet50 & 0.57s & 1.10s & 1.98s & 3.92s\\
			\hline
			3Net ResNet50 & 0.80s & 1.55s & 2.95s & 5.87s\\
			\hline
		\end{tabular}
		\caption{Run time comparison between Godard et al. \cite{godard2017unsupervised} and 3Net running single and double forward on a CPU Intel Core i7-7700K.}
		\label{tab:time}
	\end{table}
	
	\section{Runtime analysis}
	\label{sec:time}
	
	In this section, we briefly compare the runtime of 3Net compared to the models by Godard et al. \cite{godard2017unsupervised}. On high-end GPUs (e.g., Titan X Pascal), the difference between the two models either running single or double forward is negligible, taking between 0.09 and 0.11 seconds both. Nevertheless, in case of applications deploying different architectures the margin rises. 
	
	In particular, Table \ref{tab:time} compares the execution times of the considered models using ResNet50 encoder on a CPU Intel Core i7-7700K. Times are averaged on the entire Eigen split testing set. We report numbers at $256\times512$ resolution (i.e., the dimensions used by \cite{godard2017unsupervised} at inference time), as well as at full KITTI resolution, to stress how the difference between them increases with the image size. We can see how the second encoder in 3Net adds about 50\% overhead, while $2\times$ forwards usually doubles it. However, by recalling results reported in the main paper (Table 2, last 3 rows on bottom), 3Net ResNet50 running a single forward is more accurate and faster than \cite{godard2017unsupervised} ResNet50 running two forwards.

	\section*{Acknowledgements} 

	We gratefully acknowledge the support of NVIDIA Corporation with the donation of the Titan X GPU used for this research.

\end{document}